\PassOptionsToPackage{unicode}{hyperref}
\PassOptionsToPackage{hyphens}{url}
\documentclass[
  11pt,
]{article}
\usepackage{amsmath,amssymb}
\usepackage{iftex}
\ifPDFTeX
  \usepackage[T1]{fontenc}
  \usepackage[utf8]{inputenc}
  \usepackage{textcomp} 
\else 
  \usepackage{unicode-math} 
  \defaultfontfeatures{Scale=MatchLowercase}
  \defaultfontfeatures[\rmfamily]{Ligatures=TeX,Scale=1}
\fi
\usepackage{lmodern}
\ifPDFTeX\else
\fi
\IfFileExists{upquote.sty}{\usepackage{upquote}}{}
\IfFileExists{microtype.sty}{
  \usepackage[]{microtype}
  \UseMicrotypeSet[protrusion]{basicmath} 
}{}
\makeatletter
\@ifundefined{KOMAClassName}{
  \IfFileExists{parskip.sty}{%
    \usepackage{parskip}
  }{
    \setlength{\parindent}{0pt}
    \setlength{\parskip}{6pt plus 2pt minus 1pt}}
}{
  \KOMAoptions{parskip=half}}
\makeatother
\usepackage{xcolor}
\usepackage[margin=1in]{geometry}
\usepackage{longtable,booktabs,array}
\usepackage{calc} 
\usepackage{etoolbox}
\makeatletter
\patchcmd\longtable{\par}{\if@noskipsec\mbox{}\fi\par}{}{}
\makeatother
\IfFileExists{footnotehyper.sty}{\usepackage{footnotehyper}}{\usepackage{footnote}}
\makesavenoteenv{longtable}
\providecommand{\tightlist}{%
  \setlength{\itemsep}{0pt}\setlength{\parskip}{0pt}}
\ifLuaTeX
  \usepackage{selnolig}  
\fi
\IfFileExists{bookmark.sty}{\usepackage{bookmark}}{\usepackage{hyperref}}
\IfFileExists{xurl.sty}{\usepackage{xurl}}{} 
\hypersetup{
  hidelinks,
  pdfcreator={LaTeX via pandoc}}

\author{}
\date{}

\usepackage{graphicx}
\usepackage{float}
\begin{document}

\hypertarget{categorical-perception-in-large-language-model-hidden-states-structural-warping-at-digit-count-boundaries}{%
\section{Categorical Perception in Large Language Model Hidden States:
Structural Warping at Digit-Count
Boundaries}\label{categorical-perception-in-large-language-model-hidden-states-structural-warping-at-digit-count-boundaries}}

Jon-Paul Cacioli

Independent Researcher, Melbourne, Australia

Classical Minds, Modern Machines

\hypertarget{abstract}{%
\subsection{Abstract}\label{abstract}}

Categorical perception (CP), the enhanced discriminability at category
boundaries, is among the most studied phenomena in perceptual
psychology. This paper reports that analogous geometric warping occurs
in the hidden-state representations of large language models (LLMs)
processing Arabic numerals. Using representational similarity analysis
across six models from five architecture families, the study finds that
a CP-additive model (log-distance plus a boundary boost) fits the
representational geometry better than a purely continuous model at 100\%
of primary layers in every model tested. The effect is specific to
structurally defined boundaries (digit-count transitions at 10 and 100),
absent at non-boundary control positions, and absent in the temperature
domain where linguistic categories (hot/cold) lack a tokenisation
discontinuity. Two qualitatively distinct signatures emerge: ``classic
CP'' (Gemma, Qwen), where models both categorise explicitly and show
geometric warping, and ``structural CP'' (Llama, Mistral, Phi), where
geometry warps at the boundary but models cannot report the category
distinction. This dissociation is stable across boundaries and is a
property of the architecture, not the stimulus. Structural input-format
discontinuities are sufficient to produce categorical perception
geometry in LLMs, independently of explicit semantic category knowledge.

\textbf{Keywords:} categorical perception, large language models,
representational similarity analysis, numerical cognition, hidden
states, tokenisation

\hypertarget{introduction}{%
\subsection{1. Introduction}\label{introduction}}

\hypertarget{categorical-perception}{%
\subsubsection{1.1 Categorical
Perception}\label{categorical-perception}}

Categorical perception (CP) is the observation that stimuli varying
continuously along a physical dimension are perceived as falling into
discrete categories, with enhanced discriminability at category
boundaries. First demonstrated for speech sounds by Liberman, Harris,
Hoffman, and Griffith (1957), CP has since been documented across
domains including colour (Winawer et al., 2007), facial expressions
(Etcoff \& Magee, 1992), and musical pitch (Burns \& Ward, 1978). The
canonical CP signature comprises three components: (a) a sharp
identification function at the category boundary, (b) a discrimination
peak that exceeds what the identification function alone would predict,
and (c) representational warping, increased perceptual distance between
stimuli that straddle the boundary relative to equidistant
within-category pairs (Harnad, 1987; Goldstone \& Hendrickson, 2010).

Two classes of explanation dominate the literature. Under acquired CP
accounts, category learning warps the underlying perceptual space
(Goldstone, 1994). Under structural accounts, the warping reflects
properties of the representational format itself: discontinuities in the
input impose categorical structure regardless of explicit category
knowledge (McMurray, 2022). Distinguishing these accounts requires
access to the representational geometry, which is difficult in
biological systems but straightforward in artificial neural networks.

\hypertarget{digit-count-boundaries-in-numerical-cognition}{%
\subsubsection{1.2 Digit-Count Boundaries in Numerical
Cognition}\label{digit-count-boundaries-in-numerical-cognition}}

Digit-count boundaries are a fundamental property of place-value
representation. Hinrichs, Yurko, and Hu (1981) showed that reaction
times in two-digit number comparison are discontinuous at decade
boundaries. Dehaene, Dupoux, and Mehler (1990) documented both holistic
and compositional effects in multi-digit processing. Nuerk, Weger, and
Willmes (2001) identified the unit-decade compatibility effect, showing
that decades and units are processed as separate components. Poltrock
and Schwartz (1984) proposed sequential digit-by-digit comparison,
subsequently refined by evidence for parallel decomposition (Nuerk et
al., 2011).

Round numbers serve as landmarks with disproportionate frequency in
language (Dehaene \& Mehler, 1992), and the base of the number system
determines which numbers function as cognitive reference points
(Pollmann \& Jansen, 1996). Whether LLMs show similar place-value
boundary effects has received limited attention. Shah, Marupudi, Koenen,
Bhardwaj, and Varma (2023) found that LLM number representations exhibit
distance, size, and ratio effects consistent with the human mental
number line. The present study extends this work by testing whether LLM
representations show categorical structure at place-value boundaries.

\hypertarget{llms-as-a-test-case-for-structural-cp}{%
\subsubsection{1.3 LLMs as a Test Case for Structural
CP}\label{llms-as-a-test-case-for-structural-cp}}

Large language models provide a useful test case for distinguishing
structural from acquired CP. Their hidden-state representations are
directly observable, their training data are characterisable, and their
architectural properties create well-defined representational
discontinuities that can be manipulated experimentally. Arabic numerals
are a particularly clean domain: the transition from single-digit to
double-digit numbers (9 to 10) involves a simultaneous change in
character count, token count (for most tokenisers), and lexical form.
This structural discontinuity is analogous to the voice onset time
boundary in speech CP studies.

Recent work has established that LLMs encode numerical magnitude
following a logarithmic compression consistent with Weber's Law (Cacioli,
2026a). This geometry is present at all layers but is causally
implicated only at early layers, a dissociation that parallels findings
in neuroscience (Zhu et al., 2025). Park, Yun, Lee, and Shin (2024,
2025) demonstrated that LLMs represent categorical concepts as polytope
structures in hidden-state space. Bonnasse-Gahot and Nadal (2022) showed
theoretically that CP-like geometry, specifically warping of the Fisher
information metric at category boundaries, emerges in deep layers of
trained neural networks regardless of output behaviour.

The central question of this study is whether representational format
alone is sufficient to produce categorical perception geometry. If CP
geometry appears at digit-count boundaries but not at semantically
defined boundaries that lack structural discontinuities, this
constitutes evidence that format-driven warping is a valid and
independent form of categorical perception. The present study tests this
proposition by applying formal psychophysical methodology to the hidden
states of six LLMs processing Arabic numerals.

\hypertarget{the-present-study}{%
\subsubsection{1.4 The Present Study}\label{the-present-study}}

The present study tests categorical perception at digit-count boundaries
(10 and 100) in six models from five architecture families. The design
comprises five paradigms: Paradigm A (representational geometry via
RSA), Paradigm B0 (identification), Paradigm B (discrimination),
Paradigm C (precision gradient), and Paradigm E (causal intervention).
Temperature serves as a cross-domain control: a continuous physical
dimension with a linguistic category boundary (hot/cold) but no
tokenisation discontinuity. A nonce-token remapping control tests
whether the CP effect requires linguistic surface form or can be induced
by ordinal information alone.

Eight hypotheses were pre-registered on the Open Science Framework
(osf.io/qrxf3) prior to data collection, along with twelve exploratory
analyses. The hypotheses are as follows.

\begin{itemize}
\tightlist
\item
  \textbf{H0} (Falsification). CP-Additive will not outperform
  Continuous at non-boundary control positions (15, 150).
\item
  \textbf{H1} (Representational warping). CP-Additive will outperform
  Continuous at a majority of primary layers in a majority of instruct
  models.
\item
  \textbf{H2} (Behavioural discrimination). Cross-boundary pairs will
  produce higher d' than within-category pairs at matched log-distances.
\item
  \textbf{H3} (Meta-d'). Meta-d' will exceed d' at the boundary.
\item
  \textbf{H4} (Boundary contribution). Boundary-crossing will add unique
  variance beyond log-distance in hierarchical regression.
\item
  \textbf{H5} (M-ratio). Metacognitive efficiency will be higher for
  cross-boundary than within-category comparisons.
\item
  \textbf{H6} (Cross-domain). The temperature domain (hot/cold boundary,
  no tokenisation discontinuity) will show comparable CP geometry to the
  numerical domain.
\item
  \textbf{H7} (Instruction-tuning). Base and instruct variants of the
  same architecture will show comparable CP geometry.
\item
  \textbf{H8} (Identification-geometry dissociation). Some models will
  show geometric CP without explicit identification.
\end{itemize}

Pre-registered exclusion rules specified that if discrimination accuracy
reached ceiling or chance, d' would be uninformative and H2, H3, and H5
would be declared not evaluable.

The primary claim rests on representational geometry (Paradigm A),
following the precedent of Park et al.~(2024) and Bonnasse-Gahot and
Nadal (2022) who made geometric CP claims without requiring behavioural
convergence. The identification task (Paradigm B0) serves as a secondary
analysis documenting the relationship between explicit categorisation
and geometric structure. This extends the standard CP protocol by adding
a representational criterion, justified by the unique advantage of LLM
research: direct access to the representational geometry, which in human
studies must be inferred indirectly from the relationship between
identification and discrimination (McMurray, 2022).

\hypertarget{method}{%
\subsection{2. Method}\label{method}}

\hypertarget{models}{%
\subsubsection{2.1 Models}\label{models}}

Six models from five architecture families were tested, selected to
maximise architectural diversity within the constraint of fitting in 16
GB VRAM (AMD RX 7900 GRE):

\begin{longtable}[]{@{}
  >{\raggedright\arraybackslash}p{(\columnwidth - 6\tabcolsep) * \real{0.3333}}
  >{\raggedright\arraybackslash}p{(\columnwidth - 6\tabcolsep) * \real{0.1852}}
  >{\raggedright\arraybackslash}p{(\columnwidth - 6\tabcolsep) * \real{0.1852}}
  >{\raggedright\arraybackslash}p{(\columnwidth - 6\tabcolsep) * \real{0.2716}}@{}}
\toprule\noalign{}
\begin{minipage}[b]{\linewidth}\raggedright
Model
\end{minipage} & \begin{minipage}[b]{\linewidth}\raggedright
Parameters
\end{minipage} & \begin{minipage}[b]{\linewidth}\raggedright
Family
\end{minipage} & \begin{minipage}[b]{\linewidth}\raggedright
Role
\end{minipage} \\
\midrule\noalign{}
\endhead
\bottomrule\noalign{}
\endlastfoot
Llama-3-8B-Instruct & 8.0B & Meta & Primary \\
Mistral-7B-Instruct-v0.3 & 7.2B & Mistral AI & Primary \\
Gemma-2-9B-IT & 9.2B & Google & Primary \\
Qwen2.5-7B-Instruct & 7.6B & Alibaba & Primary \\
Phi-3.5-mini-instruct & 3.8B & Microsoft & Primary (scale probe) \\
Llama-3-8B-Base & 8.0B & Meta & Exploratory (instruction-tuning
control) \\
\end{longtable}

All models were loaded in FP16 (BF16 for Gemma) with
\texttt{output\_hidden\_states=True}. Phi-3.5-mini required a community
fork (Lexius/Phi-3.5-mini-instruct) with manual DynamicCache
compatibility patches for Transformers 5.0.0.

\hypertarget{stimuli}{%
\subsubsection{2.2 Stimuli}\label{stimuli}}

\hypertarget{numerical-domain-primary}{%
\paragraph{2.2.1 Numerical Domain
(Primary)}\label{numerical-domain-primary}}

\textbf{Decade-10 condition.} Seventeen probing values (4--20) spanning
the single-digit/double-digit boundary at 10. The boundary is defined
structurally by the digit-count transition, not empirically from
identification performance (see §1.1 and v0.5 revision rationale). Four
of the six models' tokenisers exhibit a token-count discontinuity at
this boundary: Gemma and Qwen encode single-digit numbers as one token
and double-digit numbers as two tokens; Mistral and Phi show the same
pattern offset by a leading-space token (2$\rightarrow$3 tokens). Llama-3 (both
instruct and base) is the exception: its BPE vocabulary includes merged
tokens for common multi-digit numbers, so both single- and double-digit
numbers are encoded as single tokens (e.g., ``9''$\rightarrow$\texttt{{[}24{]}},
``10''$\rightarrow$\texttt{{[}605{]}}). That Llama nonetheless shows strong CP
geometry despite having no token-count discontinuity is a critical
dissociation. It indicates that token-count changes amplify the boundary
effect but are not required. Character-count and lexical-form changes
are sufficient (see Supplementary Table S2 for exact token IDs across
all models). The 100-boundary involves an analogous transition
(two-digit to three-digit character strings) and a token-count increase
for Gemma, Qwen, Mistral, and Phi, producing the 3.9--12.7$\times$ effect-size
amplification reported in §3.1.1.

\textbf{Control-15 condition.} Nine probing values (11--19) centred on
15, a non-boundary position in the same numerical range. If CP-like
warping appears equally at 15 as at 10, the effect is general
representational inhomogeneity, not categorical perception (H0
falsification control).

\textbf{Decade-100 condition.} Thirteen probing values (70--130)
spanning the double-digit/triple-digit boundary at 100, with a matched
control at 150 (Control-150: values 130--170).

Eight carrier sentences embedded each probing value in natural language
contexts (e.g., ``Approximately \{N\} cases were observed''). Following
Rogers and Davis (2009), sentences were split: indices 0--3 for
identification (Paradigm B0) and indices 4--7 for RSA centroids
(Paradigm A), preventing repetition artefacts from contaminating the
geometry.

\hypertarget{temperature-domain-secondary}{%
\paragraph{2.2.2 Temperature Domain
(Secondary)}\label{temperature-domain-secondary}}

Eighteen probing values ($-$20 to 100$^{\circ}$C) spanning the hot/cold linguistic
boundary (\textasciitilde22$^{\circ}$C), with a control condition (35--51$^{\circ}$C, no
boundary). Temperature has a linguistic category distinction (hot/cold)
but no tokenisation discontinuity (``21'' and ``23'' tokenise
identically). The temperature domain tests whether linguistic category
knowledge alone, without structural input-format discontinuity, is
sufficient to induce representational CP.

\hypertarget{nonce-token-remapping-control-e10}{%
\paragraph{2.2.3 Nonce-Token Remapping Control
(E10)}\label{nonce-token-remapping-control-e10}}

Seventeen nonce tokens (``glorp'', ``blicket'', ``tazmo'', ``fenwick'',
\ldots) mapped to ordinal positions 1--17 (corresponding to values
4--20). Two conditions: \emph{nonce\_no\_order} (nonce tokens embedded
in carrier sentences with no ordering information) and
\emph{nonce\_ordered} (a preamble establishing the complete ordering
preceded each sentence). Theoretical RDMs used ordinal position as the
continuous baseline (d\_ij = \textbar rank\_i $-$ rank\_j\textbar), not
log-magnitude. This control tests whether CP requires the
linguistic/tokenisation structure of real numbers or can be induced by
ordinal information alone.

\hypertarget{paradigm-a-representational-geometry-rsa}{%
\subsubsection{2.3 Paradigm A: Representational Geometry
(RSA)}\label{paradigm-a-representational-geometry-rsa}}

For each probing value, hidden states were extracted at all layers
(embedding + transformer layers) from four RSA carrier sentences
(indices 4--7). Centroids were computed by averaging across sentences,
yielding one representation per value per layer per model.

Pairwise cosine distances between centroids formed the empirical
representational dissimilarity matrix (RDM) at each layer. Euclidean
distance was computed as a co-primary metric. Five theoretical RDMs were
compared:

\begin{enumerate}
\def\labelenumi{\arabic{enumi}.}
\tightlist
\item
  \textbf{Continuous (Weber/Log):} d\_ij = \textbar log(x\_i) $-$
  log(x\_j)\textbar{}
\item
  \textbf{CP-Additive:} d\_ij = \textbar log(x\_i) $-$ log(x\_j)\textbar{}
  + $\lambda$ $\cdot$ 1{[}different category{]} ($\lambda$ = 1.0 template)
\item
  \textbf{CP-Multiplicative:} d\_ij = \textbar log(x\_i) $-$
  log(x\_j)\textbar{} $\cdot$ (1 + $\gamma$ $\cdot$ 1{[}different category{]})
\item
  \textbf{Categorical:} d\_ij = 0 if same category, 1 if different
\item
  \textbf{Linear:} d\_ij = \textbar x\_i $-$ x\_j\textbar{}
\end{enumerate}

RSA was performed via Spearman rank correlation between empirical and
theoretical RDMs, with significance assessed by Mantel permutation tests
(10,000 permutations per layer). Layerwise multiple comparisons were
corrected using the Benjamini-Hochberg false discovery rate (FDR) at $\alpha$ =
.05. The pre-registered model comparison hierarchy was: (1) Primary:
Continuous vs CP-Additive; (2) Mechanism: CP-Additive vs
CP-Multiplicative.

\hypertarget{paradigm-b0-identification}{%
\subsubsection{2.4 Paradigm B0:
Identification}\label{paradigm-b0-identification}}

Three identification framings per model: ``small/large'',
``single-digit/multi-digit'', and ``one digit/two digits''. Each framing
used a chat-templated forced-choice format with counterbalanced A/B
option order (following Weber Appendix A methodology; Author, 2026a).
P(category\_b) was computed as the average across both orders to correct
for position bias. Sigmoid functions were fit to each identification
curve to estimate crossover point, slope, and R$^2$.

\hypertarget{paradigm-b-behavioural-discrimination}{%
\subsubsection{2.5 Paradigm B: Behavioural
Discrimination}\label{paradigm-b-behavioural-discrimination}}

Forced-choice discrimination (``Which is larger: A or B?'') was
administered for 900 trials spanning cross-boundary and within-category
pairs at six log-distance levels, with counterbalanced option order. The
primary dependent variable was \textbar $\Delta$logit\textbar{} (decision
confidence), used as a reaction-time analogue: models process all tokens
in a single forward pass, so confidence magnitude serves as a
computational proxy for processing ease (Cacioli, 2026b). Full trial
structure is reported in the Supplementary Materials.

\begin{figure}[H]
\centering
\includegraphics[width=\textwidth]{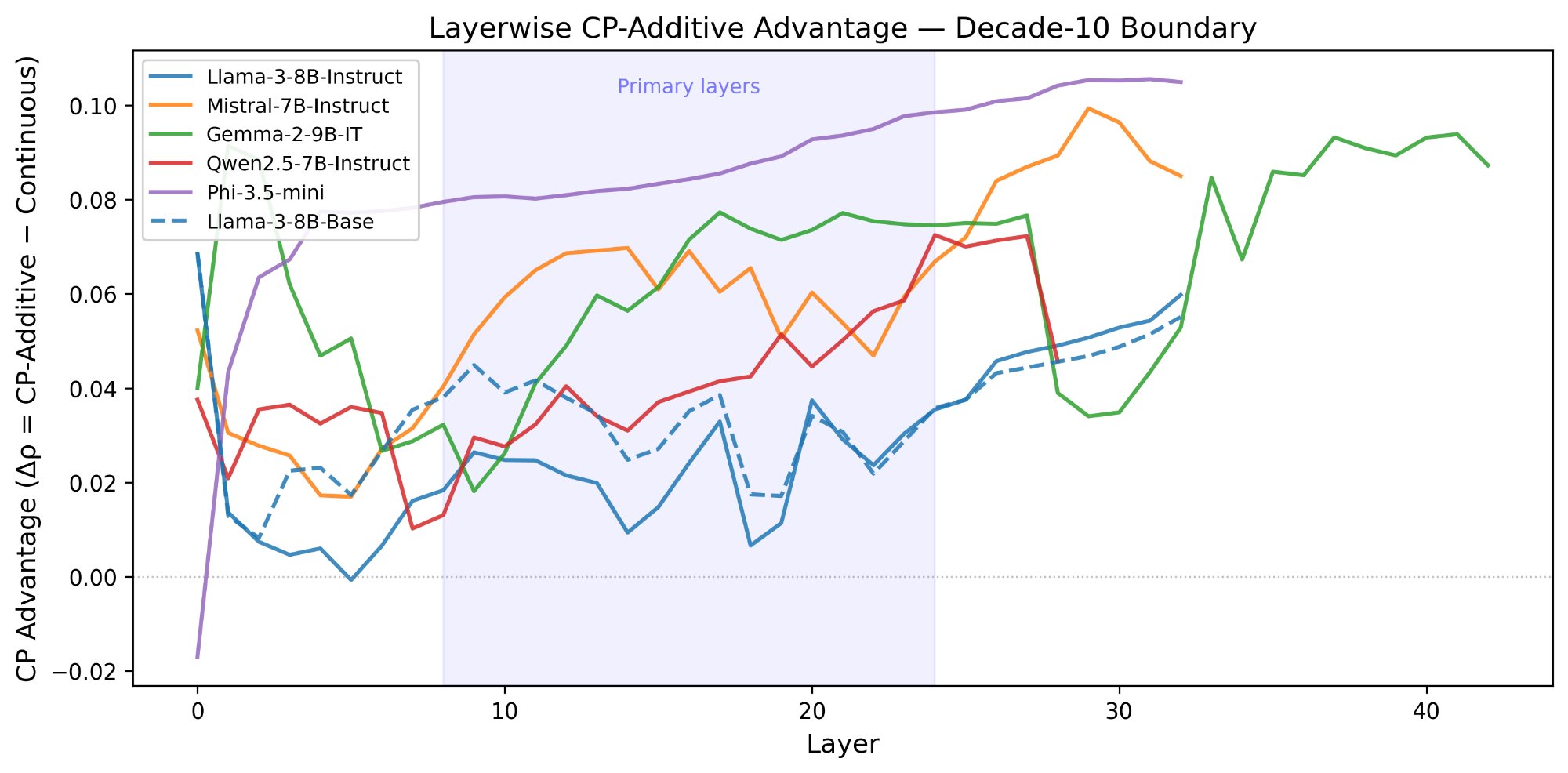}
\caption{Layerwise CP-Additive advantage ($\Delta\rho$) for six models at the decade-10 boundary. Shaded region: primary layers. Llama-3-8B-Base (dashed) shows comparable geometry.}
\label{fig:layerwise}
\end{figure}

\hypertarget{paradigm-c-precision-gradient}{%
\subsubsection{2.6 Paradigm C: Precision
Gradient}\label{paradigm-c-precision-gradient}}

Local representational precision was defined as
1/\textbar\textbar h(n+1) $-$ h(n)\textbar\textbar{} for adjacent probing
values at each layer, following Cacioli, J.-P. (2026a). CP predicts a distance
spike (precision dip) at the category boundary.

\hypertarget{paradigm-e-causal-intervention-e5}{%
\subsubsection{2.7 Paradigm E: Causal Intervention
(E5)}\label{paradigm-e-causal-intervention-e5}}

A ridge-regression probe was trained on Paradigm A centroids predicting
binary category membership (\textless{} 10 vs $\geq$ 10), defining the
``category direction'' at each layer. Direction validity was confirmed
via PCA: PC1 correlation with category membership (mean $\rho$ = .83 at
primary layers) and probe-PC1 cosine similarity.

Activation patching was performed via PyTorch forward hooks: at each
target layer, hidden states were modified in-place by adding $\alpha$ $\cdot$
\textbar\textbar w\textbar\textbar{} $\cdot$ v\_cat, where v\_cat is the unit
category direction, \textbar\textbar w\textbar\textbar{} is the probe
weight norm, and $\alpha$ $\in$ \{0.25, 0.50, 0.75, 1.00\}. Ten random-direction
controls (same norm) established the specificity baseline. The dependent
variable was $\Delta$(\textbar $\Delta$logit\textbar), the change in discrimination
confidence under patching.

\hypertarget{h4-boundary-contribution-to-representational-distance}{%
\subsubsection{2.8 H4: Boundary Contribution to Representational
Distance}\label{h4-boundary-contribution-to-representational-distance}}

Hierarchical regression at each primary layer tested whether
boundary-crossing adds unique variance beyond log-distance. Step 1:
regress empirical pairwise distances on \textbar log(x\_i) $-$
log(x\_j)\textbar. Step 2: add a boundary-crossing indicator (1 if pair
straddles the boundary, 0 otherwise). $\Delta$R$^2$ and F-test for the addition of
the boundary predictor.

\hypertarget{pre-registration-and-statistical-analysis}{%
\subsubsection{2.9 Pre-Registration and Statistical
Analysis}\label{pre-registration-and-statistical-analysis}}

All hypotheses and analysis pipelines were pre-registered on OSF
(osf.io/qrxf3) prior to data collection. Confirmatory tests used
Bonferroni correction within hypothesis families ($\alpha$ = .05). Effect sizes
(Cohen's d, $\Delta$R$^2$, Spearman $\rho$) are reported for all comparisons. Bootstrap
95\% confidence intervals (10,000 resamples, seed = 42) are reported
where applicable. All code, stimuli, and analysis scripts are available
at https://github.com/synthiumjp/weber (directory
\texttt{m3\_pilot/}). Seed = 42 throughout.

Pre-registered exclusion rules: If discrimination accuracy is at ceiling
($\geq$ 99\%) or chance ($\leq$ 51\%), d' is uninformative and hypotheses
contingent on d' (H2, H2b, H3, H5) are declared not evaluable. If no
identification framing produces a crossover within the probing range,
the McMurray (2022) predicted-vs-observed test (H2b) is not evaluable.

\hypertarget{results}{%
\subsection{3. Results}\label{results}}

Results are organised by paradigm, following the analysis hierarchy
specified in the pre-registration. Section 3.1 presents the primary
confirmatory analysis (representational geometry). Sections 3.2--3.4
present secondary analyses (identification, discrimination, and
precision gradient). Sections 3.5--3.7 present the three control
conditions (non-boundary control, temperature domain, and nonce-token
remapping). Section 3.8 presents the causal intervention. Section 3.9
summarises hypothesis outcomes.

\hypertarget{representational-geometry-cp-additive-dominates-at-all-layers-h1}{%
\subsubsection{3.1 Representational Geometry: CP-Additive Dominates at
All Layers
(H1)}\label{representational-geometry-cp-additive-dominates-at-all-layers-h1}}

The primary confirmatory test compared five theoretical RDMs against
empirical representational dissimilarity matrices at each layer of each
model. Table 1 summarises the decade-10 condition
(single-digit/double-digit boundary at 10).

\textbf{Table 1.} Paradigm A results: decade-10 boundary. CP-Additive vs
Continuous comparison across six models. ``CP \textgreater{} Cont''
reports the number of primary layers where the CP-Additive model
achieved higher Spearman $\rho$ than the Continuous model. ``Mean $\Delta$$\rho$'' is the
mean $\rho$ difference across primary layers. All Mantel permutation tests
(10,000 permutations) were significant at p \textless{} .001 at all
primary layers after Benjamini-Hochberg FDR correction at $\alpha$ = .05.

\begin{longtable}[]{@{}
  >{\raggedright\arraybackslash}p{(\columnwidth - 10\tabcolsep) * \real{0.2785}}
  >{\raggedright\arraybackslash}p{(\columnwidth - 10\tabcolsep) * \real{0.1392}}
  >{\raggedright\arraybackslash}p{(\columnwidth - 10\tabcolsep) * \real{0.1392}}
  >{\raggedright\arraybackslash}p{(\columnwidth - 10\tabcolsep) * \real{0.1392}}
  >{\raggedright\arraybackslash}p{(\columnwidth - 10\tabcolsep) * \real{0.1392}}
  >{\raggedright\arraybackslash}p{(\columnwidth - 10\tabcolsep) * \real{0.1392}}@{}}
\toprule\noalign{}
\begin{minipage}[b]{\linewidth}\raggedright
Model
\end{minipage} & \begin{minipage}[b]{\linewidth}\raggedright
Primary Layers
\end{minipage} & \begin{minipage}[b]{\linewidth}\raggedright
CP \textgreater{} Cont
\end{minipage} & \begin{minipage}[b]{\linewidth}\raggedright
Mean $\Delta$$\rho$
\end{minipage} & \begin{minipage}[b]{\linewidth}\raggedright
Max $\rho$ (CP-Add)
\end{minipage} & \begin{minipage}[b]{\linewidth}\raggedright
Max $\rho$ (Cont)
\end{minipage} \\
\midrule\noalign{}
\endhead
\bottomrule\noalign{}
\endlastfoot
Llama-3-8B-Instruct & 17/17 & 17/17 & +0.023 & 0.940 & 0.921 \\
Mistral-7B-Instruct & 17/17 & 17/17 & +0.060 & 0.929 & 0.889 \\
Gemma-2-9B-IT & 23/23 & 23/23 & +0.063 & 0.890 & 0.834 \\
Qwen2.5-7B-Instruct & 15/15 & 15/15 & +0.035 & 0.832 & 0.788 \\
Phi-3.5-mini & 17/17 & 17/17 & +0.087 & 0.698 & 0.600 \\
Llama-3-8B-Base & 17/17 & 17/17 & +0.032 & 0.956 & 0.927 \\
\end{longtable}

CP-Additive achieved higher Spearman correlation with the empirical RDM
than the Continuous model at 100\% of primary layers for all six models
(H1 supported). The pre-registered success criterion ($\geq$50\% of primary
layers for $\geq$3/5 instruct models) was exceeded maximally. Figure 1 shows
the empirical RDM for Llama-3-8B-Instruct at layer 16, illustrating the
block structure at the 9/10 boundary. The CP-Additive advantage is not a
trivial consequence of adding a free parameter: the boundary boost $\lambda$ was
fixed at 1.0 (not fit to data), the effect was absent at non-boundary
control positions (§3.5.1), absent in the temperature domain (§3.5.2),
and the hierarchical regression (§3.1.2) confirmed that
boundary-crossing contributes 5--27\% unique variance beyond
log-distance. This is a non-trivial geometric feature, not an artefact
of model complexity. The effect was robust to distance metric: Euclidean
distance produced identical results (CP-Additive \textgreater{}
Continuous at 100\% of primary layers, all models; Supplementary Table
S1). Euclidean max $\rho$ for the CP-Additive model ranged from 0.702 (Phi)
to 0.955 (Base).

\begin{figure}[H]
\centering
\includegraphics[width=0.7\textwidth]{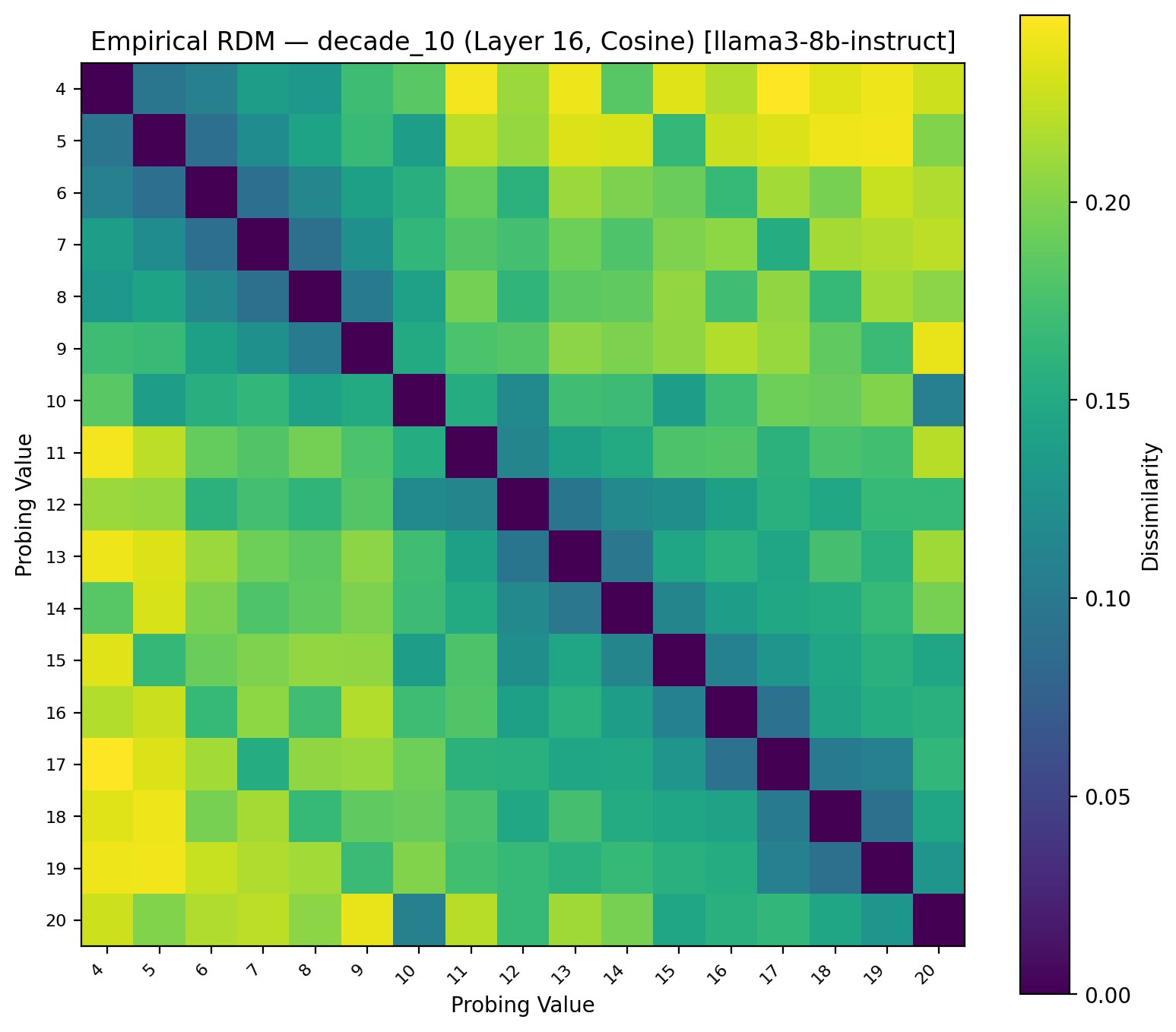}
\caption{Empirical RDM for Llama-3-8B-Instruct at layer 16 (decade-10, cosine distance). Block structure at the 9/10 boundary visible as increased cross-boundary distances.}
\label{fig:rdm}
\end{figure}

The pre-registered mechanism test compared CP-Additive (constant
boundary boost) with CP-Multiplicative (proportional scaling).
CP-Additive provided superior fit at a majority of primary layers in
five of six models: Mistral (17/17), Gemma (22/23), Qwen (13/15), Phi
(17/17), and Base (12/17). Llama-Instruct was split (8/17). This
establishes that the boundary effect operates as a constant additive
displacement in representational space rather than a proportional
scaling of existing distances, consistent with the Fisher information
warping predicted by Bonnasse-Gahot and Nadal (2022).

\hypertarget{replication-at-the-100-boundary}{%
\paragraph{3.1.1 Replication at the
100-Boundary}\label{replication-at-the-100-boundary}}

The decade-100 condition (double-digit/triple-digit boundary at 100,
probing values 70--130) replicated the decade-10 pattern (Table 1b).
CP-Additive exceeded Continuous at 100\% of primary layers for all six
models, with substantially larger effect sizes.

\textbf{Table 1b.} Paradigm A results: decade-100 boundary and
control-150. Same format as Table 1. E4 ratio = decade-100 mean CP
advantage / decade-10 mean CP advantage.

\begin{longtable}[]{@{}
  >{\raggedright\arraybackslash}p{(\columnwidth - 12\tabcolsep) * \real{0.2651}}
  >{\raggedright\arraybackslash}p{(\columnwidth - 12\tabcolsep) * \real{0.0964}}
  >{\raggedright\arraybackslash}p{(\columnwidth - 12\tabcolsep) * \real{0.1084}}
  >{\raggedright\arraybackslash}p{(\columnwidth - 12\tabcolsep) * \real{0.1325}}
  >{\raggedright\arraybackslash}p{(\columnwidth - 12\tabcolsep) * \real{0.1084}}
  >{\raggedright\arraybackslash}p{(\columnwidth - 12\tabcolsep) * \real{0.1325}}
  >{\raggedright\arraybackslash}p{(\columnwidth - 12\tabcolsep) * \real{0.1325}}@{}}
\toprule\noalign{}
\begin{minipage}[b]{\linewidth}\raggedright
Model
\end{minipage} & \begin{minipage}[b]{\linewidth}\raggedright
CP \textgreater{} Cont (100)
\end{minipage} & \begin{minipage}[b]{\linewidth}\raggedright
Mean $\Delta$$\rho$ (100)
\end{minipage} & \begin{minipage}[b]{\linewidth}\raggedright
Max $\rho$ (CP-Add)
\end{minipage} & \begin{minipage}[b]{\linewidth}\raggedright
Max $\rho$ (Cont)
\end{minipage} & \begin{minipage}[b]{\linewidth}\raggedright
Ctrl-150 $\Delta$$\rho$
\end{minipage} & \begin{minipage}[b]{\linewidth}\raggedright
E4 Ratio
\end{minipage} \\
\midrule\noalign{}
\endhead
\bottomrule\noalign{}
\endlastfoot
Llama-3-8B-Instruct & 17/17 & +0.319 & 0.676 & 0.442 & +0.017 & 12.7$\times$ \\
Mistral-7B-Instruct & 17/17 & +0.268 & 0.323 & 0.136 & +0.042 & 3.9$\times$ \\
Gemma-2-9B-IT & 23/23 & +0.279 & 0.394 & 0.260 & +0.016 & 4.0$\times$ \\
Qwen2.5-7B-Instruct & 15/15 & +0.162 & 0.404 & 0.241 & +0.012 & 4.6$\times$ \\
Phi-3.5-mini & 17/17 & +0.476 & 0.580 & 0.177 & +0.012 & 4.6$\times$ \\
Llama-3-8B-Base & 17/17 & +0.305 & 0.619 & 0.338 & +0.016 & 8.8$\times$ \\
\end{longtable}

The 100-boundary effect was 3.9--12.7$\times$ larger than the decade-10 effect
across models (E4). The matched control at 150 (probing values 130--170)
showed negligible CP advantage ($\Delta$$\rho$ = +0.012 to +0.042), far smaller than
both the decade-100 and decade-10 effects.

The scaling of effect size with boundary magnitude is interpretable: the
100-boundary involves a transition from two-digit to three-digit
numbers, which changes both character count and token count for all
tested tokenisers, producing a larger structural discontinuity than the
10-boundary.

\hypertarget{hierarchical-regression-h4}{%
\paragraph{3.1.2 Hierarchical Regression
(H4)}\label{hierarchical-regression-h4}}

At each primary layer, pairwise representational distances were
regressed on log-distance (Step 1), then a boundary-crossing indicator
was added (Step 2). The boundary predictor added significant unique
variance at 100\% of primary layers for all six models (all F-test p
\textless{} .001). Table 2 summarises the hierarchical regression
results.

\textbf{Table 2.} H4 hierarchical regression: unique variance
contributed by boundary-crossing beyond log-distance.

\begin{longtable}[]{@{}
  >{\raggedright\arraybackslash}p{(\columnwidth - 6\tabcolsep) * \real{0.2933}}
  >{\raggedright\arraybackslash}p{(\columnwidth - 6\tabcolsep) * \real{0.2267}}
  >{\raggedright\arraybackslash}p{(\columnwidth - 6\tabcolsep) * \real{0.2267}}
  >{\raggedright\arraybackslash}p{(\columnwidth - 6\tabcolsep) * \real{0.2267}}@{}}
\toprule\noalign{}
\begin{minipage}[b]{\linewidth}\raggedright
Model
\end{minipage} & \begin{minipage}[b]{\linewidth}\raggedright
Sig layers
\end{minipage} & \begin{minipage}[b]{\linewidth}\raggedright
Mean $\Delta$R$^2$
\end{minipage} & \begin{minipage}[b]{\linewidth}\raggedright
Max $\Delta$R$^2$
\end{minipage} \\
\midrule\noalign{}
\endhead
\bottomrule\noalign{}
\endlastfoot
Llama-3-8B-Instruct & 17/17 & 0.050 & 0.084 \\
Mistral-7B-Instruct & 17/17 & 0.145 & 0.200 \\
Gemma-2-9B-IT & 23/23 & 0.174 & 0.231 \\
Qwen2.5-7B-Instruct & 15/15 & 0.079 & 0.115 \\
Phi-3.5-mini & 17/17 & 0.267 & 0.331 \\
Llama-3-8B-Base & 17/17 & 0.067 & 0.113 \\
\end{longtable}

Boundary-crossing explains 5--27\% of variance in representational
distance beyond what log-magnitude compression accounts for. The
proportionally largest $\Delta$R$^2$ appears in Phi (0.267), which has the weakest
overall geometry (max $\rho$ = 0.747). The boundary effect is proportionally
dominant in the smaller model even though its absolute geometry is
attenuated.

\hypertarget{identification-classic-cp-vs-structural-cp-h8}{%
\subsubsection{3.2 Identification: Classic CP vs Structural CP
(H8)}\label{identification-classic-cp-vs-structural-cp-h8}}

Three identification framings (``small/large'',
``single-digit/multi-digit'', ``one digit/two digits'') were
administered to each instruct model using counterbalanced A/B
forced-choice. Table 3 summarises identification outcomes using the
digit\_count framing, which produced the best signal across models.

\textbf{Table 3.} Paradigm B0 results: identification at the decade-10
boundary. ``Boundary at 10?'' indicates whether the identification
function crossed 0.50 within $\pm$1 step of the structural boundary.

\begin{longtable}[]{@{}
  >{\raggedright\arraybackslash}p{(\columnwidth - 6\tabcolsep) * \real{0.2821}}
  >{\raggedright\arraybackslash}p{(\columnwidth - 6\tabcolsep) * \real{0.2051}}
  >{\raggedright\arraybackslash}p{(\columnwidth - 6\tabcolsep) * \real{0.2436}}
  >{\raggedright\arraybackslash}p{(\columnwidth - 6\tabcolsep) * \real{0.2436}}@{}}
\toprule\noalign{}
\begin{minipage}[b]{\linewidth}\raggedright
Model
\end{minipage} & \begin{minipage}[b]{\linewidth}\raggedright
Boundary at 10?
\end{minipage} & \begin{minipage}[b]{\linewidth}\raggedright
Pattern
\end{minipage} & \begin{minipage}[b]{\linewidth}\raggedright
CP Type
\end{minipage} \\
\midrule\noalign{}
\endhead
\bottomrule\noalign{}
\endlastfoot
Llama-3-8B-Instruct & No & Gradient, never crosses 0.50 & Structural
CP \\
Mistral-7B-Instruct & No & Extreme position bias; $\approx$0.50 after
counterbalancing & Structural CP \\
Gemma-2-9B-IT & \textbf{Yes} & Sharp step at exactly 10 (slope = 6.3) &
Classic CP \\
Qwen2.5-7B-Instruct & \textbf{Yes} & Step at 11--12 (shifted 1--2 steps
from structural boundary) & Classic CP \\
Phi-3.5-mini & No & Flat at $\approx$0.50; no category signal & Structural CP \\
Llama-3-8B-Base & N/A & No chat template; raw-logit identification
uninformative & Structural CP \\
\end{longtable}

Two qualitatively distinct patterns emerged (Figure 3). Gemma and Qwen
showed \emph{classic CP}: both geometric warping and explicit
categorisation at the boundary, with sharp sigmoid identification
functions. Llama, Mistral, Phi, and the base model showed
\emph{structural CP}: geometric warping without explicit identification.
The category boundary is imposed by representational format rather than
explicit category knowledge.

\begin{figure}[H]
\centering
\includegraphics[width=\textwidth]{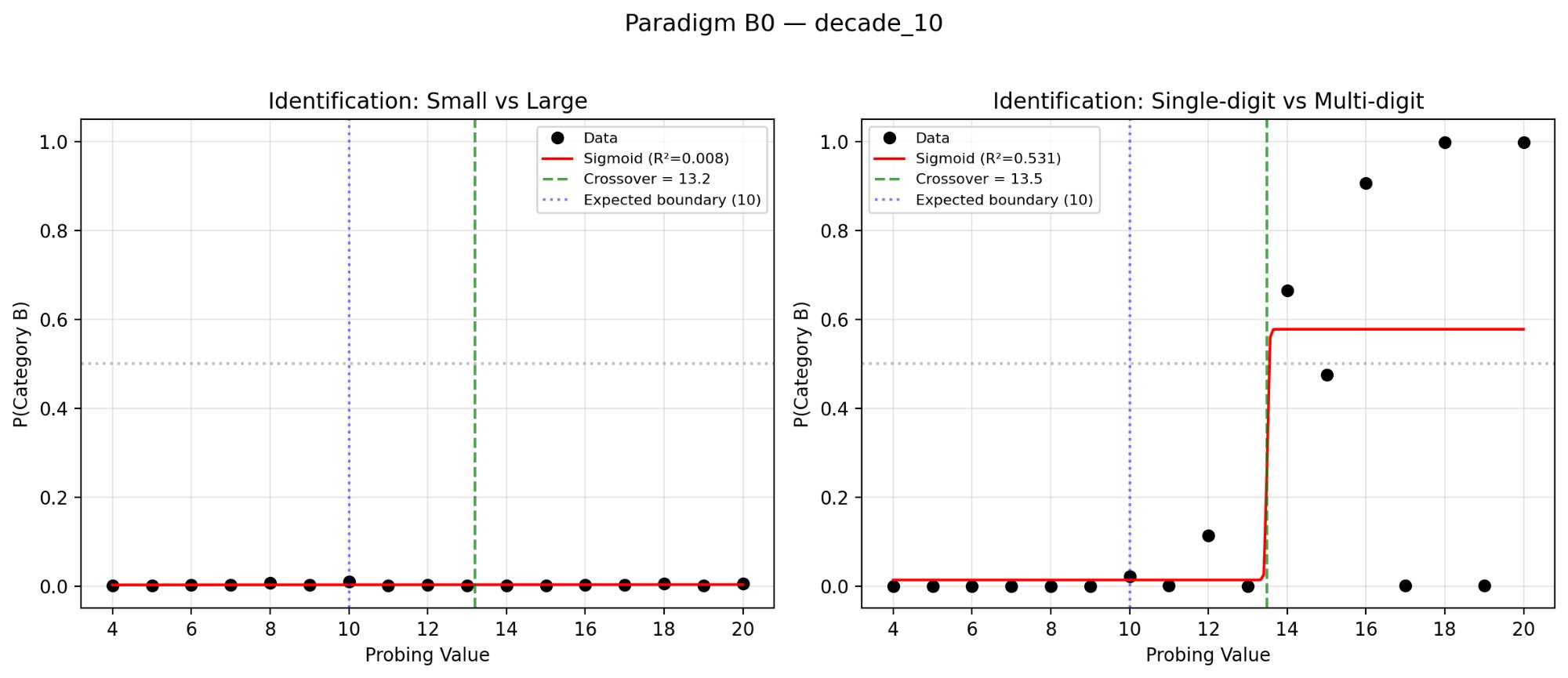}
\includegraphics[width=\textwidth]{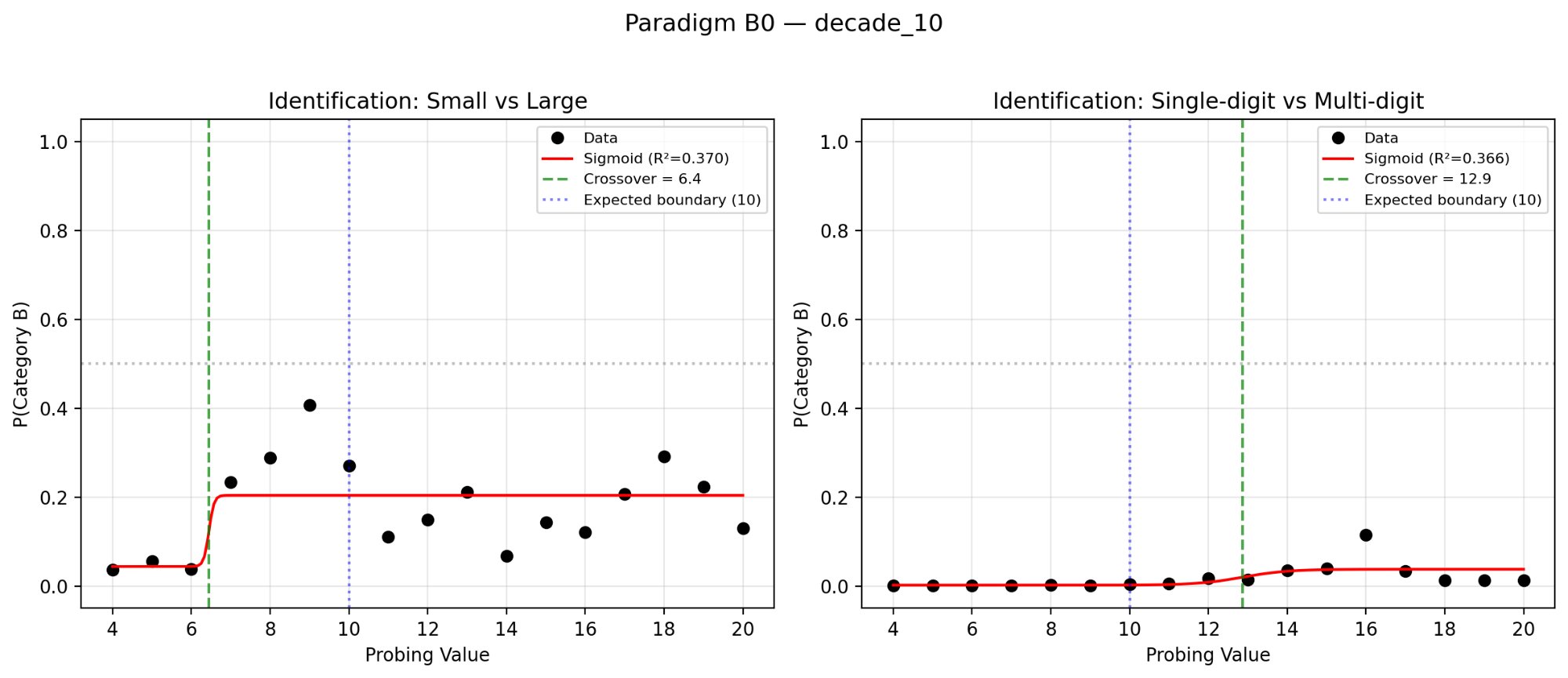}
\caption{Identification functions at decade-10 boundary. Top: Gemma (classic CP, sharp step). Bottom: Llama (structural CP, flat). Both show comparable geometry (Table 1).}
\label{fig:identification}
\end{figure}

This dissociation was stable across boundaries: at the 100-boundary, the
same models that failed to identify the 10-boundary also failed to
identify the 100-boundary, and vice versa. The dissociation is therefore
a property of the architecture, not the stimulus.

Cross-model correlation confirmed the dissociation: identification slope
did not predict geometric CP strength (Spearman $\rho$ = $-$0.14, p = .79, n =
6; E9). Models with the sharpest identification boundaries (Gemma: slope
= 6.34) did not show stronger geometric CP than models with no
identification at all (Phi: slope = 0.0, yet largest $\Delta$R$^2$ = 0.331 in the
hierarchical regression). This confirms that geometric CP and explicit
identification are dissociated, consistent with H8.

Prompt robustness analysis (E6) revealed that only Qwen produced
consistent crossover locations across all three framings (within 0.51
steps). Gemma and Phi showed framing-dependent boundaries (\textgreater2
steps disagreement), and Llama, Mistral, and the base model produced no
crossover under any framing.

\hypertarget{behavioural-discrimination-h2}{%
\subsubsection{3.3 Behavioural Discrimination
(H2)}\label{behavioural-discrimination-h2}}

\hypertarget{accuracy}{%
\paragraph{3.3.1 Accuracy}\label{accuracy}}

All four capable instruct models (Llama, Mistral, Gemma, Qwen) achieved
$\geq$99.9\% accuracy on the forced-choice ``Which is larger?''
discrimination task. Phi-3.5-mini and Llama-3-8B-Base showed
chance-level accuracy after counterbalancing (pure position bias). All
instruct models except Phi exhibited strong first-option (A) bias prior
to counterbalancing; Phi showed second-option (B) bias, an
architecture-dependent effect that extends the position bias documented
in Paradigm B0 (E12). Per the pre-registered exclusion rule, d$^{\prime}$ was
uninformative and H2 (accuracy-based), H2b (McMurray strict test), H3
(meta-d$^{\prime}$), and H5 (M-ratio) were declared not evaluable.

\hypertarget{confidence-rt-analogue}{%
\paragraph{3.3.2 Confidence (RT
Analogue)}\label{confidence-rt-analogue}}

Although accuracy was at ceiling, the magnitude of the decision signal,
\textbar $\Delta$logit\textbar{} between the chosen and unchosen option,
varied systematically with boundary position (cf.~Author, 2026b, for the
use of \textbar $\Delta$logit\textbar{} as a confidence measure in LLM
psychophysics). This is an operational proxy for the processing-ease
signal measured by reaction time in human CP studies, not a behavioural
equivalent; its validity rests on the assumption that larger
representational distance at the decision layer produces larger logit
separation, which is confirmed by the Paradigm A geometry.
Cross-boundary pairs produced significantly higher confidence than
within-category pairs at matched log-distances (Table 4).

\textbf{Table 4.} Paradigm B results: confidence-based discrimination.
$\Delta$Conf = mean \textbar $\Delta$logit\textbar{} for cross-boundary minus
within-category pairs. Cohen's d and Mann-Whitney U test for the
confidence difference. ``Sig levels'' reports the number of log-distance
bins (of 6) where the confidence difference reached significance at p
\textless{} .05.

\begin{longtable}[]{@{}
  >{\raggedright\arraybackslash}p{(\columnwidth - 12\tabcolsep) * \real{0.2418}}
  >{\raggedright\arraybackslash}p{(\columnwidth - 12\tabcolsep) * \real{0.1538}}
  >{\raggedright\arraybackslash}p{(\columnwidth - 12\tabcolsep) * \real{0.1648}}
  >{\raggedright\arraybackslash}p{(\columnwidth - 12\tabcolsep) * \real{0.1099}}
  >{\raggedright\arraybackslash}p{(\columnwidth - 12\tabcolsep) * \real{0.1099}}
  >{\raggedright\arraybackslash}p{(\columnwidth - 12\tabcolsep) * \real{0.0989}}
  >{\raggedright\arraybackslash}p{(\columnwidth - 12\tabcolsep) * \real{0.0989}}@{}}
\toprule\noalign{}
\begin{minipage}[b]{\linewidth}\raggedright
Model
\end{minipage} & \begin{minipage}[b]{\linewidth}\raggedright
Conf(cross)
\end{minipage} & \begin{minipage}[b]{\linewidth}\raggedright
Conf(within)
\end{minipage} & \begin{minipage}[b]{\linewidth}\raggedright
$\Delta$Conf
\end{minipage} & \begin{minipage}[b]{\linewidth}\raggedright
Cohen's d
\end{minipage} & \begin{minipage}[b]{\linewidth}\raggedright
p
\end{minipage} & \begin{minipage}[b]{\linewidth}\raggedright
Sig levels
\end{minipage} \\
\midrule\noalign{}
\endhead
\bottomrule\noalign{}
\endlastfoot
Llama-3-8B-Instruct & 8.39 & 7.77 & +0.62 & 0.45 & \textless.001 &
4/6 \\
Mistral-7B-Instruct & 10.59 & 10.16 & +0.43 & 0.29 & \textless.001 &
3/6 \\
Gemma-2-9B-IT & 8.63 & 8.44 & +0.18 & 0.47 & \textless.001 & 4/6 \\
Qwen2.5-7B-Instruct & 20.94 & 20.70 & +0.23 & 0.23 & .001 & 2/6 \\
Phi-3.5-mini & 0.005 & 0.005 & +0.0003 & 0.09 & .38 & 1/6 \\
Llama-3-8B-Base & 0.075 & 0.061 & +0.015 & 0.55 & \textless.001 & 4/6 \\
\end{longtable}

The confidence effect was significant for five of six models (all except
Phi), with medium effect sizes (Cohen's d = 0.23--0.55). This
demonstrates that the geometric warping documented in Paradigm A
propagates to behavioural output: the additive boundary boost in
representational distance translates to larger evidence magnitude at the
decision stage.

The distance-controlled pattern was interpretable: the confidence
difference was significant at larger log-distances (0.30--0.60) but not
at the smallest distances (0.10--0.20), where pairs are close together
regardless of boundary position.

Llama-3-8B-Base showed the largest effect size (d = 0.55) despite pure
position bias at the accuracy level: the magnitude of its
position-biased logit varied with boundary position, consistent with the
Paradigm A finding that base and instruct models share categorical
geometric structure.

Phi-3.5-mini showed no confidence effect (d = 0.09, p = .38), despite
exhibiting geometric CP in Paradigm A (CP-Additive \textgreater{}
Continuous at 17/17 primary layers). This dissociation between geometry
and behavioural propagation further supports the geometry-function
distinction: the additive boundary boost is present in the
representational space of the smaller model but is insufficient,
given Phi's attenuated overall geometry (max $\rho$ = 0.698) and near-zero
confidence magnitudes (\textbar $\Delta$logit\textbar{} $\approx$ 0.005), to propagate
to the output layer as a measurable confidence difference.

\hypertarget{precision-gradient-paradigm-c}{%
\subsubsection{3.4 Precision Gradient (Paradigm
C)}\label{precision-gradient-paradigm-c}}

Local representational precision (1/\textbar\textbar h(n+1) $-$
h(n)\textbar\textbar) showed a boundary-specific spike at 9$\rightarrow$10 for all
six models. Precision ratios (boundary distance / mean non-boundary
distance) ranged from 1.42 (Phi) to 2.29 (Mistral) at the decade-10
boundary, compared with ratios near 1.0 (0.92--1.07) at the matched
control position 15. This local expansion of representational distance
at the boundary corresponds directly to the Fisher information peak
predicted by Bonnasse-Gahot and Nadal (2022): in their framework,
optimal categorisation warps the metric tensor of the representational
space such that Fisher information is maximal at category boundaries.
The precision gradient measured here is an empirical proxy for this
metric warping. It quantifies the local expansion of representational
space relative to the input space at the boundary, exactly as their
theory predicts.

\hypertarget{control-conditions}{%
\subsubsection{3.5 Control Conditions}\label{control-conditions}}

\hypertarget{analytic-invariance}{%
\paragraph{3.5.0 Analytic Invariance}\label{analytic-invariance}}

Both geometric and behavioural CP measures used in this study are
analytically invariant to the sampling temperature parameter: hidden
states are computed deterministically from fixed inputs (temperature
affects only the softmax distribution over next-token probabilities),
and \textbar $\Delta$logit\textbar{} is defined on raw logits which are
unaffected by temperature scaling (E2).

\hypertarget{non-boundary-control-h0}{%
\paragraph{3.5.1 Non-Boundary Control
(H0)}\label{non-boundary-control-h0}}

At the control position (15), the Continuous model consistently
outperformed CP-Additive (negative CP advantage for all six models). The
warping is specific to structurally defined boundaries and absent at
arbitrary non-boundary positions within the same numerical range (H0 not
falsified). At control position 150, CP advantages were negligible
(+0.012 to +0.042; see Table 1b), far smaller than the decade-100
effects (+0.162 to +0.476) and comparable in magnitude to the
temperature domain negatives, confirming that the massive 100-boundary
effects are boundary-specific.

\hypertarget{temperature-domain-h6}{%
\paragraph{3.5.2 Temperature Domain (H6)}\label{temperature-domain-h6}}

The temperature domain ($-$20 to 100$^{\circ}$C, hot/cold linguistic boundary at
\textasciitilde22$^{\circ}$C) showed negative CP advantage for all six models
(Table 5b). The Continuous model outperformed CP-Additive at the
hot/cold boundary, and the temperature control condition (boundary at
43$^{\circ}$C, no linguistic distinction) showed equally negative advantages.

\textbf{Table 5b.} Temperature domain RSA results (H6). CP advantage is
the mean $\Delta$$\rho$ (CP-Additive $-$ Continuous) across primary layers. Negative
values indicate Continuous fits better.

\begin{longtable}[]{@{}
  >{\raggedright\arraybackslash}p{(\columnwidth - 6\tabcolsep) * \real{0.2933}}
  >{\raggedright\arraybackslash}p{(\columnwidth - 6\tabcolsep) * \real{0.2267}}
  >{\raggedright\arraybackslash}p{(\columnwidth - 6\tabcolsep) * \real{0.2267}}
  >{\raggedright\arraybackslash}p{(\columnwidth - 6\tabcolsep) * \real{0.2267}}@{}}
\toprule\noalign{}
\begin{minipage}[b]{\linewidth}\raggedright
Model
\end{minipage} & \begin{minipage}[b]{\linewidth}\raggedright
Hot/cold CP wins
\end{minipage} & \begin{minipage}[b]{\linewidth}\raggedright
Hot/cold $\Delta$$\rho$
\end{minipage} & \begin{minipage}[b]{\linewidth}\raggedright
Temp ctrl $\Delta$$\rho$
\end{minipage} \\
\midrule\noalign{}
\endhead
\bottomrule\noalign{}
\endlastfoot
Llama-3-8B-Instruct & 0/17 & $-$0.062 & $-$0.062 \\
Mistral-7B-Instruct & 1/17 & $-$0.038 & $-$0.061 \\
Gemma-2-9B-IT & 1/23 & $-$0.027 & $-$0.054 \\
Qwen2.5-7B-Instruct & 4/15 & $-$0.009 & $-$0.063 \\
Phi-3.5-mini & 0/17 & $-$0.039 & $-$0.055 \\
Llama-3-8B-Base & 0/17 & $-$0.069 & $-$0.065 \\
\end{longtable}

Despite having a linguistic category distinction, temperature lacks a
tokenisation discontinuity (``21'' and ``23'' tokenise identically) and
therefore does not produce representational warping. This is a key
positive null: it provides strong evidence that structural input-format
discontinuity is sufficient for CP geometry in LLMs, and that linguistic
category knowledge alone is not sufficient in this setting (H6 not
supported).

\hypertarget{nonce-token-remapping-control-e10-1}{%
\paragraph{3.5.3 Nonce-Token Remapping Control
(E10)}\label{nonce-token-remapping-control-e10-1}}

The nonce-token experiment tested whether categorical geometry can be
induced by ordinal information alone, without the linguistic surface
form of real numbers. Seventeen nonce tokens (``glorp'', ``blicket'',
``tazmo'', \ldots) were mapped to ordinal positions 1--17 in two
conditions: no ordering information (nonce\_no\_order) and an explicit
preamble establishing the ordering (nonce\_ordered). Table 5 summarises
E10 results.

\textbf{Table 5.} E10 nonce-token remapping control: CP-Additive
advantage across conditions and models.

\begin{longtable}[]{@{}
  >{\raggedright\arraybackslash}p{(\columnwidth - 6\tabcolsep) * \real{0.2933}}
  >{\raggedright\arraybackslash}p{(\columnwidth - 6\tabcolsep) * \real{0.2267}}
  >{\raggedright\arraybackslash}p{(\columnwidth - 6\tabcolsep) * \real{0.2267}}
  >{\raggedright\arraybackslash}p{(\columnwidth - 6\tabcolsep) * \real{0.2267}}@{}}
\toprule\noalign{}
\begin{minipage}[b]{\linewidth}\raggedright
Model
\end{minipage} & \begin{minipage}[b]{\linewidth}\raggedright
nonce\_no\_order $\Delta$$\rho$
\end{minipage} & \begin{minipage}[b]{\linewidth}\raggedright
nonce\_ordered $\Delta$$\rho$
\end{minipage} & \begin{minipage}[b]{\linewidth}\raggedright
decade\_10 $\Delta$$\rho$
\end{minipage} \\
\midrule\noalign{}
\endhead
\bottomrule\noalign{}
\endlastfoot
Llama-3-8B-Instruct & $-$0.010 & +0.012 & +0.023 \\
Mistral-7B-Instruct & $-$0.002 & +0.005 & +0.060 \\
Gemma-2-9B-IT & +0.004 & +0.004 & +0.063 \\
Qwen2.5-7B-Instruct & $-$0.010 & +0.017 & +0.035 \\
Phi-3.5-mini & $-$0.023 & +0.007 & +0.087 \\
Llama-3-8B-Base & $-$0.008 & +0.004 & +0.032 \\
\end{longtable}

\textbf{nonce\_no\_order:} Clean null. No significant geometry and no CP
at the arbitrary boundary position. Mean CP advantage was negative or
negligible for all models ($-$0.023 to +0.004). The control works as
designed.

\textbf{nonce\_ordered:} Ordinal geometry emerged strongly, with a small
but consistent CP advantage at the boundary (mean +0.004 to +0.017
across models). However, this CP effect was 3--10$\times$ smaller than the
decade-10 effect and 10--100$\times$ smaller than the decade-100 effect.

E10 establishes a three-level hierarchy (Figure 5). Without ordering
information, nonce tokens produce no geometry and no CP. With ordering
information, models build ordinal representations from context alone,
with a weak CP-like effect at the boundary position. With real numbers,
the tokenisation and digit-count discontinuity amplifies the boundary
effect by an order of magnitude. The linguistic surface form is the
amplifier, not the sole cause.

\begin{figure}[H]
\centering
\includegraphics[width=\textwidth]{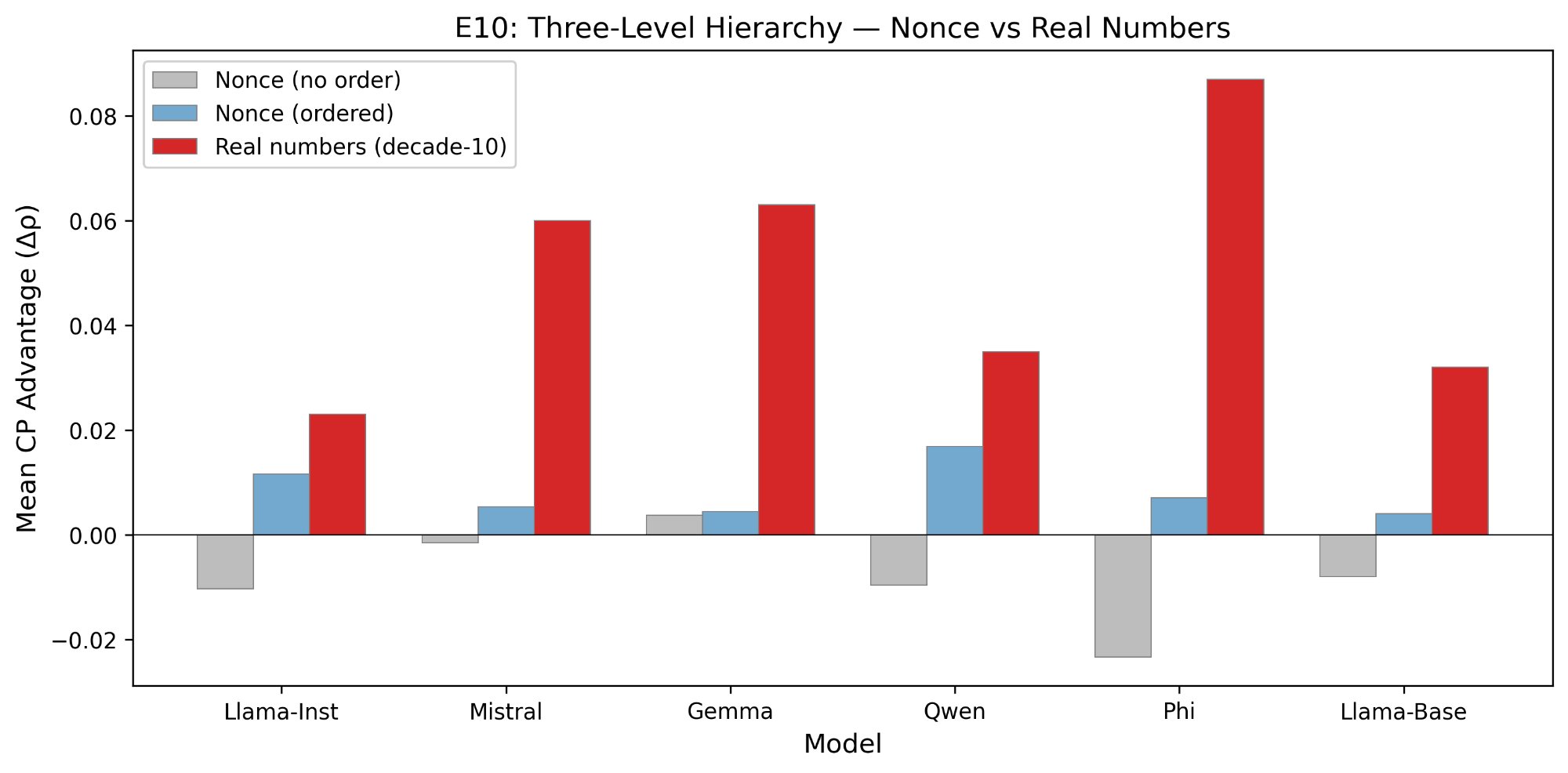}
\caption{E10 nonce-token control: three-level hierarchy. Mean CP advantage ($\Delta\rho$) for nonce (no order), nonce (ordered), and real numbers. Real numbers produce 3--10$\times$ larger effects.}
\label{fig:nonce}
\end{figure}

A notable case is Phi-3.5-mini, which achieved near-perfect ordinal
geometry in the nonce\_ordered condition (max $\rho$ = 0.97 for the
continuous model) yet its CP advantage was only +0.007. This
dissociation between geometry strength and CP magnitude demonstrates
that strong ordinal geometry does not automatically produce categorical
warping; the warping requires a structural discontinuity.

\hypertarget{instruction-tuning-control-h7}{%
\subsubsection{3.6 Instruction-Tuning Control
(H7)}\label{instruction-tuning-control-h7}}

Llama-3-8B-Base showed nearly identical CP geometry to
Llama-3-8B-Instruct (CP advantage +0.032 vs +0.023; precision ratio 1.71
vs 1.57). The base model achieved the highest max $\rho$ in the study
(CP-Additive $\rho$ = 0.955), confirming that categorical geometric structure
is present in pretrained representations and is not introduced by
instruction tuning (H7 supported).

Where base and instruct models diverge is in identification and
discrimination behaviour: the base model cannot categorise or compare
numbers in forced-choice tasks (no chat template, pure position bias at
the accuracy level), yet its representational geometry contains the same
categorical signature. This is a dissociation between representational
structure and behavioural competence.

\hypertarget{layerwise-profile-e1-and-local-manifold-analysis-e7-e8}{%
\subsubsection{3.7 Layerwise Profile (E1) and Local Manifold Analysis
(E7, E8)}\label{layerwise-profile-e1-and-local-manifold-analysis-e7-e8}}

CP geometry (the CP-Additive advantage over Continuous) emerged at
primary layers and was absent at early and late layers for all models
(E1; Figure 2). The boundary position induced a local manifold rotation:
the first principal component of the local neighbourhood rotated
81.6--89.6$^{\circ}$ at the decade-10 boundary relative to non-boundary positions
(E7; range across models: Mistral 81.6$^{\circ}$, Gemma 82.3$^{\circ}$, Qwen 86.5$^{\circ}$,
Llama-Base 88.8$^{\circ}$, Llama-Instruct 89.1$^{\circ}$, Phi 89.6$^{\circ}$). Phase-reset analysis
confirmed that representational similarity trajectories showed a
significant discontinuity at the boundary (Mann-Whitney p \textless{}
.01, all models; mean ratio of boundary-to-non-boundary distance:
1.51--2.37; E8). The relationship between CP strength ($\lambda$) and global
compression quality ($\beta$) varied across models: larger models (Gemma,
Mistral, Llama-Base) showed anticorrelation at the decade-100 boundary
($\rho$ = $-$0.30 to $-$0.79), suggesting that CP locally disrupts the continuous
magnitude geometry, while Phi showed strong positive correlation ($\rho$ =
+0.88 to +0.99), consistent with the boundary effect being the dominant
geometric feature in the smaller model (E11).

\hypertarget{causal-intervention-e5}{%
\subsubsection{3.8 Causal Intervention
(E5)}\label{causal-intervention-e5}}

Activation patching tested whether the categorical geometry identified
in Paradigm A is causally implicated in discrimination behaviour. This
analysis was performed on Llama-3-8B-Instruct as a proof-of-concept;
full cross-architecture replication is left for future work. A
ridge-regression probe trained on RSA centroids (probe accuracy = 1.00
at all primary layers; PC1--category correlation: mean $\rho$ = .83) defined
the ``category direction'' at each layer. Patching along this direction
at graded dose levels ($\alpha$ $\in$ \{0.25, 0.50, 0.75, 1.00\}) produced large,
specific, dose-dependent changes in discrimination confidence at early
layers but negligible effects at mid-to-late layers where CP geometry is
strongest (Table 6; Figure 4).

\textbf{Table 6.} E5 causal intervention results (Llama-3-8B-Instruct).
\textbar $\Delta$conf\textbar{} = absolute change in discrimination confidence
under category-direction patching. Specificity = ratio of
category-direction effect to mean random-direction effect (10 random
controls).

\begin{longtable}[]{@{}
  >{\raggedright\arraybackslash}p{(\columnwidth - 10\tabcolsep) * \real{0.1410}}
  >{\raggedright\arraybackslash}p{(\columnwidth - 10\tabcolsep) * \real{0.2051}}
  >{\raggedright\arraybackslash}p{(\columnwidth - 10\tabcolsep) * \real{0.1410}}
  >{\raggedright\arraybackslash}p{(\columnwidth - 10\tabcolsep) * \real{0.1410}}
  >{\raggedright\arraybackslash}p{(\columnwidth - 10\tabcolsep) * \real{0.1410}}
  >{\raggedright\arraybackslash}p{(\columnwidth - 10\tabcolsep) * \real{0.2051}}@{}}
\toprule\noalign{}
\begin{minipage}[b]{\linewidth}\raggedright
Layer
\end{minipage} & \begin{minipage}[b]{\linewidth}\raggedright
Depth
\end{minipage} & \begin{minipage}[b]{\linewidth}\raggedright
Cat
\end{minipage} & \begin{minipage}[b]{\linewidth}\raggedright
$\Delta$conf
\end{minipage} & \begin{minipage}[b]{\linewidth}\raggedright
\end{minipage} & \begin{minipage}[b]{\linewidth}\raggedright
Rand
\end{minipage} \\
\midrule\noalign{}
\endhead
\bottomrule\noalign{}
\endlastfoot
5 & Early & 0.621 & 0.009 & 70.1$\times$ & Monotonic \\
8 & Early-primary & 0.242 & 0.006 & 43.6$\times$ & Monotonic \\
16 & Mid-primary & 0.018 & 0.002 & 10.8$\times$ & Monotonic \\
23 & Near peak RSA & 0.058 & 0.002 & 34.3$\times$ & Monotonic \\
27 & Late & 0.010 & 0.001 & 12.1$\times$ & Non-monotonic \\
\end{longtable}

\begin{figure}[H]
\centering
\includegraphics[width=\textwidth]{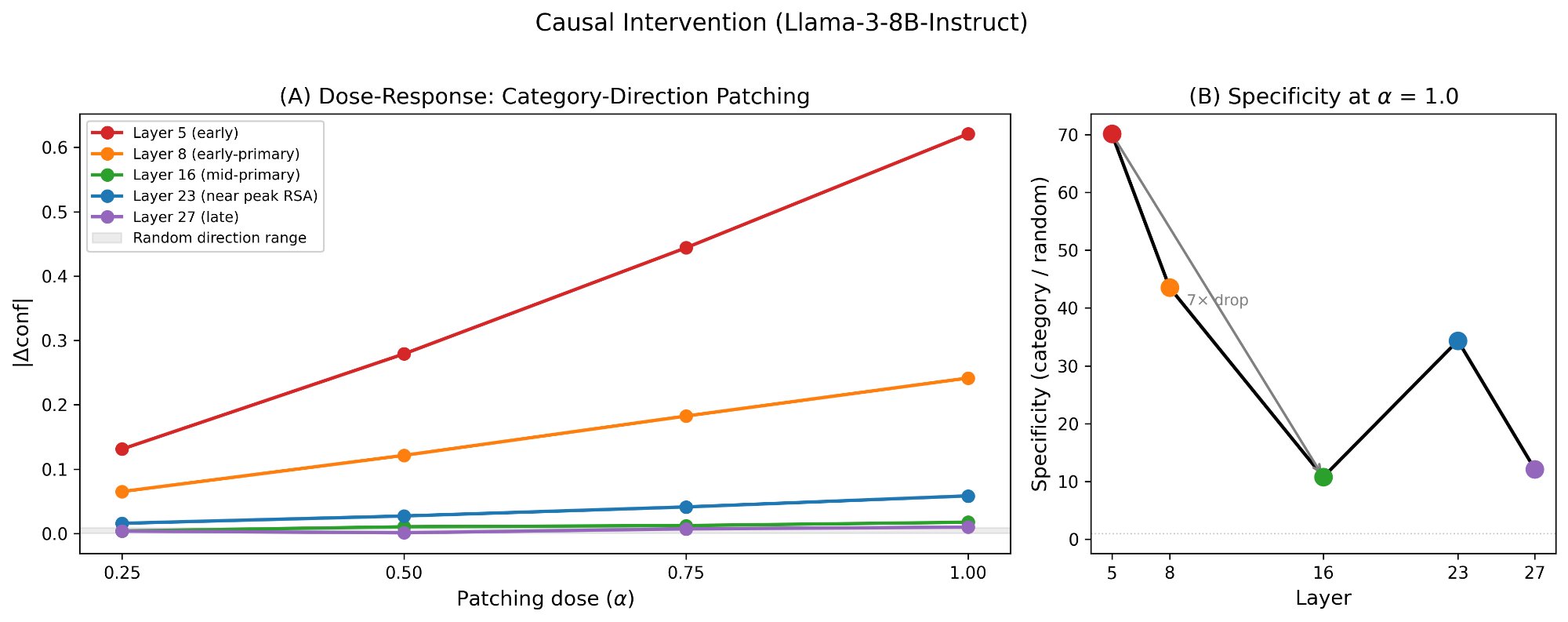}
\caption{Causal intervention (E5, Llama-3-8B-Instruct). (A) Dose-response: $|\Delta\mathrm{conf}|$ vs patching dose. Early layers show strong effects; late layers flat. (B) Specificity at $\alpha = 1.0$.}
\label{fig:causal}
\end{figure}

The absolute effect of category-direction patching dropped by
approximately 60$\times$ from layer 5 (\textbar $\Delta$conf\textbar{} = 0.621) to
layer 16 (\textbar $\Delta$conf\textbar{} = 0.018). The category direction at
early layers, where geometry is still forming, produced large, specific,
dose-dependent changes in discrimination confidence. At mid-to-late
layers where CP geometry peaks, patching produced negligible effects
despite remaining direction-specific (specificity 10--34$\times$). This pattern
replicates exactly the dissociation found in the Weber study (Cacioli,
2026a) for magnitude representations: early layers are causally
implicated, late peak-RSA layers are causally inert.

A sign reversal across layers was observed: layer 5 patching increased
confidence (+0.621), while layer 8 patching decreased it ($-$0.242). Both
effects were monotonic and highly specific, indicating that the category
direction has opposite functional roles at different depths.

This suggests a programme-level principle: representational structure
and functional relevance dissociate across depth. The layer with the
most interpretable geometry is not the layer performing the
computational work. This holds for both magnitude (Cacioli, 2026a) and
category (present study).

\hypertarget{hypothesis-summary}{%
\subsubsection{3.9 Hypothesis Summary}\label{hypothesis-summary}}

Table 7 summarises outcomes for all pre-registered hypotheses.

\textbf{Table 7.} Pre-registered hypothesis outcomes.

\begin{longtable}[]{@{}
  >{\raggedright\arraybackslash}p{(\columnwidth - 4\tabcolsep) * \real{0.3243}}
  >{\raggedright\arraybackslash}p{(\columnwidth - 4\tabcolsep) * \real{0.3243}}
  >{\raggedright\arraybackslash}p{(\columnwidth - 4\tabcolsep) * \real{0.3243}}@{}}
\toprule\noalign{}
\begin{minipage}[b]{\linewidth}\raggedright
Hypothesis
\end{minipage} & \begin{minipage}[b]{\linewidth}\raggedright
Outcome
\end{minipage} & \begin{minipage}[b]{\linewidth}\raggedright
Key Evidence
\end{minipage} \\
\midrule\noalign{}
\endhead
\bottomrule\noalign{}
\endlastfoot
H0 (Falsification) & Not falsified & Control positions (15, 150) show no
CP \\
H1 (Representational warping) & \textbf{Supported} & CP-Add
\textgreater{} Cont at 100\% of primary layers, all 6 models \\
H2 (Behavioural d$^{\prime}$) & Not evaluable & Ceiling accuracy (pre-registered
exclusion) \\
H2b (McMurray strict test) & Not evaluable & No clean identification
boundary in most models \\
H3 (Meta-d$^{\prime}$) & Not evaluable & Ceiling accuracy (pre-registered
exclusion) \\
H4 (Boundary contribution) & \textbf{Supported} & $\Delta$R$^2$ = 0.05--0.27 at
100\% of layers, all models \\
H5 (M-ratio) & Not evaluable & Ceiling accuracy (pre-registered
exclusion) \\
H6 (Cross-domain) & \textbf{Not supported} & Temperature shows no CP
(positive null) \\
H7 (Instruction-tuning) & \textbf{Supported} & Base $\approx$ Instruct for
geometry; diverge for behaviour \\
H8 (ID-geometry dissociation) & \textbf{Supported} & Gemma/Qwen =
classic CP; Llama/Mistral/Phi = structural CP \\
\end{longtable}

Four hypotheses were supported (H1, H4, H7, H8). One hypothesis yielded
an informative negative (H6). Four hypotheses were not evaluable due to
pre-registered exclusion criteria (ceiling accuracy in the
discrimination task). All twelve pre-registered exploratory analyses
were completed.

\hypertarget{discussion}{%
\subsection{4. Discussion}\label{discussion}}

The present study applied formal psychophysical methodology (representational similarity analysis, signal detection theory,
counterbalanced forced-choice psychophysics, and causal intervention)
to the hidden states of six large language models processing Arabic
numerals. The results establish that categorical perception, one of the
most extensively studied phenomena in perceptual psychology, has a
structural analogue in artificial neural networks. This section
considers what the findings mean for theories of categorical perception,
what they reveal about the relationship between representational
structure and behavioural competence, and what methodological
implications they carry for the study of both biological and artificial
cognition.

\hypertarget{categorical-perception-without-category-knowledge}{%
\subsubsection{4.1 Categorical Perception Without Category
Knowledge}\label{categorical-perception-without-category-knowledge}}

The central finding is that all six models show categorical warping at
digit-count boundaries. CP-Additive outperforms Continuous at 100\% of
primary layers, with the effect scaling by 4--13$\times$ at the larger
boundary, yet only two of six models can explicitly identify the
category distinction when asked. This dissociation between
representational structure and explicit categorisation (H8) is the most
theoretically consequential result.

In the human CP literature, the relationship between identification and
discrimination has been treated as definitional. McMurray (2022) argued
forcefully that many putative demonstrations of CP fail his strict test:
observed discrimination must exceed what is predicted from the
identification function alone. The logic of this test presupposes that
the only evidence for categorical structure is behavioural, because in
human studies, the representational geometry is not directly observable.

LLMs extend this framework. Direct access to the geometry reveals that
categorical warping is present in models that cannot report the
category. The four ``structural CP'' models (Llama, Mistral, Phi, Base)
exhibit geometric warping at the boundary without any ability to perform
the identification task. This is not a failure of the CP paradigm. It is
a dissociation that McMurray's framework, designed for systems where
representations are opaque, was not positioned to detect.

The implication for cognitive science is that the standard two-task
criterion (identification + discrimination) is sufficient but not
necessary for establishing CP. When representational geometry is
directly accessible, geometric warping at the boundary is itself
evidence for categorical structure, regardless of whether the system can
report the category. This aligns with the theoretical positions of
Bonnasse-Gahot and Nadal (2022), who showed that Fisher information
warping at category boundaries emerges in deep layers of trained neural
networks regardless of output behaviour, and Park et al.~(2024), who
demonstrated categorical polytope structure without requiring
identification tasks. The present study provides the first empirical
demonstration that this representational-geometric form of CP appears
alongside, and dissociates from, the classical behavioural form within
the same set of systems.

\hypertarget{structural-vs-acquired-categorical-perception}{%
\subsubsection{4.2 Structural vs Acquired Categorical
Perception}\label{structural-vs-acquired-categorical-perception}}

The distinction between structural and acquired CP has been difficult to
adjudicate in biological systems because the two sources of warping are
typically confounded. Humans who learn colour categories also have
retinal cone tuning curves; humans who learn phoneme categories also
have cochlear filter banks. Input structure and learned categorisation
covary.

LLMs offer a cleaner separation. Three converging results establish that
the CP documented here is primarily structural rather than acquired.

First, the temperature domain control (H6). Temperature has a linguistic
category boundary (hot/cold) but no tokenisation discontinuity. ``21$^{\circ}$C''
and ``23$^{\circ}$C'' tokenise identically. If CP were driven primarily by
learned semantic categories, temperature should show at least some
warping. It does not. CP advantage was negative for all six models in
the temperature domain ($-$0.009 to $-$0.069), compared with positive
advantages of +0.023 to +0.087 for numbers. This provides strong
evidence that structural input-format discontinuity is sufficient for CP
geometry in this setting. The negative result should be interpreted
conservatively: ``hot/cold'' is a relatively fuzzy, context-dependent
boundary, and a sharper linguistic category (e.g., grammatical number)
would provide a more demanding test of whether semantic categories can
ever produce CP in LLMs.

Second, the nonce-token remapping control (E10). Nonce tokens mapped to
ordinal positions with an explicit ordering preamble produced a weak CP
effect ($\Delta$$\rho$ = +0.004 to +0.017), but this was 3--10$\times$ smaller than the
decade-10 effect and 10--100$\times$ smaller than the decade-100 effect.
Ordinal context alone can induce a trace of boundary warping, likely
reflecting the model's ability to learn weak categorical structure from
in-context ordinal information, but the tokenisation and digit-count
discontinuity amplifies this effect by an order of magnitude. The Phi
dissociation is particularly informative: near-perfect ordinal geometry
($\rho$ = 0.97) with negligible CP advantage (+0.007), demonstrating that
ordered representations do not automatically produce categorical
warping. Strong ordinal geometry is not sufficient; the structural
discontinuity in the input format appears necessary for the large
effects documented for real numbers.

Third, the instruction-tuning control (H7). Llama-3-8B-Base and
Llama-3-8B-Instruct show nearly identical CP geometry ($\Delta$$\rho$ = +0.032 vs
+0.023; the base model actually achieves the highest max $\rho$ in the
study). The categorical structure is present in pretrained
representations before any instruction tuning. It is a property of how
the model encodes the distributional statistics of digit strings, not a
product of explicit category instruction.

Together, these three controls converge on the same conclusion:
structural properties of the input (tokenisation boundaries, digit-count
transitions, character-length changes) are sufficient to produce
categorical perception geometry in LLMs, and the effect does not require
learned semantic category knowledge. This does not mean that all CP is
structural, nor that semantic categories could never produce CP under
different conditions. The classic/structural CP dissociation (H8)
suggests that instruction tuning can add an explicit categorisation
capacity on top of the structural geometry, producing the full CP
signature in some architectures (Gemma, Qwen) but not others. The
structural warping, however, is universal.

This result has implications beyond LLMs. It provides a proof of concept
that categorical perception can arise from representational format
constraints without category learning. This possibility has been
theorised (McMurray, 2022; Massaro, 1987) but never demonstrated in a
system where the geometry, the training data, and the input format are
all simultaneously observable. This does not imply that human CP is
primarily structural, only that such a mechanism is computationally
sufficient in a learning system.

\hypertarget{the-geometry-function-dissociation}{%
\subsubsection{4.3 The Geometry-Function
Dissociation}\label{the-geometry-function-dissociation}}

The causal intervention (E5) revealed a dissociation between
representational structure and functional relevance that replicates
across both the magnitude (Cacioli, 2026a) and category domains. At early
layers (5 and 8), patching along the category direction produced large,
specific, dose-dependent changes in discrimination confidence
(\textbar $\Delta$conf\textbar{} up to 0.621, specificity 44--70$\times$ above random
controls). At mid-to-late layers where CP geometry peaks (layers
16--27), patching produced negligible effects (\textbar $\Delta$conf\textbar{}
= 0.01--0.06) despite remaining direction-specific.

This pattern challenges the common interpretability practice of
declaring the layer with the strongest probe accuracy the
``representation layer'' for a given concept. The layer with the most
interpretable geometry is not the layer performing the computation. The
geometry is real and is causally grounded at early layers, but the
computational work happens where the geometry is still forming, not
where it has crystallised into its most legible form.

The sign reversal at layers 5 and 8 (positive and negative effects on
confidence, respectively) further suggests that the category direction
plays functionally different roles at different depths. This is
consistent with recent work on superposition and polysemanticity in
neural network representations (Elhage et al., 2022): the same geometric
feature may support different computations at different layers, and a
single ``category direction'' extracted by a linear probe may conflate
functionally distinct subspaces.

For the CP literature, this finding complicates the use of
representational similarity analysis as a standalone measure of
categorical structure. RSA can identify where CP geometry is present,
but not where it matters. Causal methods are required to establish
functional relevance. This lesson applies equally to neuroimaging
studies of human categorical perception, where the distinction between
representational presence and causal relevance is often elided.

\hypertarget{additive-vs-multiplicative-warping}{%
\subsubsection{4.4 Additive vs Multiplicative
Warping}\label{additive-vs-multiplicative-warping}}

The pre-registered mechanism test established that the boundary effect
is additive rather than multiplicative. The boundary produces a constant
displacement in representational distance, not a proportional scaling of
existing distances. Five of six models favoured the additive model; only
Llama-Instruct was split.

This finding connects to the theoretical framework of Bonnasse-Gahot and
Nadal (2022), who derived that optimal categorisation produces a local
spike in Fisher information at the category boundary. This manifests
geometrically as a local additive increase in representational distance.
The additive nature of the effect is important because it means the
boundary operates as an independent structural feature layered on top of
the continuous magnitude geometry, rather than modulating that geometry.
This is consistent with the hierarchical regression results (H4), where
boundary-crossing explained 5--27\% of unique variance beyond
log-distance: the boundary adds information to the geometry rather than
distorting the information already present.

\hypertarget{why-transformers-produce-boundary-warping}{%
\subsubsection{4.5 Why Transformers Produce Boundary
Warping}\label{why-transformers-produce-boundary-warping}}

The results are consistent with the following mechanistic account of how
structural CP arises in transformer architectures. At the input stage,
digit-count boundaries produce a discontinuity in the embedding space.
Single-digit numbers map to one token (or one character); double-digit
numbers map to two. This creates a discrete jump in the input
representation that is independent of numerical magnitude. The embedding
layer encodes this discontinuity directly: representations of ``9'' and
``10'' differ not only in magnitude but in format, producing an initial
separation in embedding space.

Subsequent transformer layers preserve and amplify this separation
through two processes. First, self-attention operates over token
sequences whose length and composition change at the boundary, producing
qualitatively different attention patterns for single-digit and
double-digit inputs even when the semantic content (magnitude) varies
smoothly. Second, the residual stream accumulates these format-dependent
features across layers, producing the additive displacement documented
in the RSA results. The additive rather than multiplicative nature of
the effect is consistent with this account: the boundary introduces a
constant offset in the representation, independent of the magnitude
distance between stimuli.

This mechanistic account is consistent with three converging
observations. The Llama dissociation shows that token-count
discontinuity is not required but character-count discontinuity is
sufficient, implicating the earliest stages of input processing. The
instruction-tuning control shows that the effect is present in
pretrained representations, indicating that it is acquired from the
distributional statistics of digit strings during pretraining rather
than from explicit category instruction. The layerwise profile shows
that CP geometry strengthens across depth, consistent with amplification
through successive transformer layers.

This account does not explain why some architectures (Gemma, Qwen)
additionally develop explicit identification behaviour. A plausible
hypothesis is that instruction tuning reinforces the boundary signal
sufficiently for it to influence the output distribution, but this
remains to be tested.

\hypertarget{implications-for-theories-of-categorical-perception}{%
\subsubsection{4.6 Implications for Theories of Categorical
Perception}\label{implications-for-theories-of-categorical-perception}}

The present study sits at the intersection of classical psychophysics,
computational neuroscience, and AI interpretability. Three sets of
implications follow.

\emph{From cognitive science to AI:} The psychophysical toolkit provides
a framework for characterising how neural networks represent categorical
structure that goes beyond the binary ``does the model know X?''
question typical of NLP evaluation. Instead, it asks how knowledge is
geometrically encoded, how it varies across depth, and whether
representational structure is causally linked to behaviour.

\emph{From AI to cognitive science:} LLMs provide a system where the
acquired-vs-structural distinction can be cleanly tested because the
representational format, the training data, and the resulting geometry
are all observable. The finding that structural input properties
(tokenisation, digit count) produce categorical warping without explicit
category learning offers an existence proof for the structural account
of CP that has been debated for decades in the human literature
(Massaro, 1987; McMurray, 2022). It does not resolve the debate for
biological systems, but it establishes that the structural mechanism is
computationally viable. Format-driven warping can produce the full
geometric CP signature in a learning system.

The findings also connect to the numerical cognition literature on
place-value processing. The decade boundary effects documented by
Hinrichs et al.~(1981) and Dehaene et al.~(1990) in human reaction times
have a representational counterpart in LLM hidden states: the additive
boundary boost at the 9/10 and 99/100 transitions corresponds to the RT
discontinuities at decade boundaries that these studies identified. The
3.9 to 12.7 times amplification of CP geometry at the 100-boundary
relative to the 10-boundary parallels the finding that larger
place-value transitions produce stronger boundary effects in human
processing (Nuerk et al., 2011).

The instruction-tuning control (H7) is also relevant. The presence of
categorical geometry in pretrained representations before instruction
tuning suggests that the boundary structure is acquired from
distributional statistics of digit strings during pretraining. This is
consistent with Dehaene and Mehler's (1992) observation that round
numbers have disproportionate frequency in natural language.

Shah et al.~(2023) showed that LLM number representations exhibit
distance, size, and ratio effects consistent with the human mental
number line. The present study extends their approach by showing that
LLM representations also exhibit boundary-specific warping at
place-value transitions, a phenomenon that has not previously been
tested in artificial systems using the formal CP paradigm.

The present work complements a growing body of research applying
cognitive science constructs to LLM representations. Park et al.~(2024,
2025) demonstrated that LLMs represent categorical concepts as polytope
structures, establishing that categorical geometry exists in hidden
states but without testing the psychophysical signature (identification,
discrimination, boundary-specific warping) that defines CP. Shani,
Marjieh, and Griffiths (2025) showed that LLM representations compress
along typicality gradients in a manner consistent with human
categorisation, but focused on within-category structure rather than the
between-category boundary effects that are diagnostic of CP. The Weber
study (Cacioli, 2026a) established that numerical magnitude follows
logarithmic compression consistent with Weber's Law, providing the
continuous baseline geometry on which the present categorical effects
are superimposed. The present study integrates these strands: it tests
CP specifically, the boundary phenomenon, using the formal
psychophysical methodology (RSA with theoretical model comparison,
SDT-based discrimination, identification functions, causal intervention)
that the prior work did not employ. The result is that LLM categorical
structure is not merely present as a geometric fact but operates through
a specific mechanism (additive boundary warping driven by input-format
discontinuity) that can be dissociated from explicit category knowledge,
quantified at each layer, and causally tested.

\hypertarget{bidirectional-implications}{%
\paragraph{Bidirectional
Implications}\label{bidirectional-implications}}

The geometry-function dissociation identifies a methodological hazard
shared by cognitive science and AI interpretability: the most legible
representational structure may not be the most functionally relevant (de
Wit et al., 2016). The present study provides a clean demonstration
because the entire computational pipeline is observable and causally
manipulable.

\hypertarget{implications-for-llm-behaviour}{%
\subsubsection{4.7 Implications for LLM
Behaviour}\label{implications-for-llm-behaviour}}

The structural CP documented here has practical consequences for LLM
numerical reasoning. If representational geometry warps at digit-count
boundaries, models may process numbers near boundaries differently from
numbers far from boundaries, even when the numerical magnitudes are
similar. This could contribute to the well-documented finding that LLMs
make more errors on arithmetic problems that cross digit-count
boundaries (e.g., 9 + 3 vs 7 + 2) and to tokenisation-induced biases in
numerical tasks. The confidence results (Table 4) provide direct
evidence that the geometric warping propagates to the output layer:
cross-boundary pairs produce larger decision signals than
within-category pairs at matched distances, meaning that the boundary
affects not just internal representations but observable model
behaviour.

More broadly, the finding that representational format shapes the
geometry of LLM hidden states, independently of semantic content,
suggests that tokenisation design is a more consequential architectural
choice than is commonly appreciated. Different tokenisers produce
different structural boundaries, and these boundaries produce measurable
geometric effects that persist across the full depth of the network.

\hypertarget{limitations}{%
\subsubsection{4.8 Limitations}\label{limitations}}

\emph{Causal generality.} The causal intervention was performed on one
model only (Llama-3-8B-Instruct). The geometry-function dissociation
replicates the pattern found in the Weber study for magnitude
representations, but the generality of the causal finding across
architectures remains to be tested.

\emph{Domain specificity.} The numerical domain is a particularly clean
case for structural CP because digit-count boundaries produce a
simultaneous change in tokenisation, character count, and lexical form.
Whether structural CP extends to domains with subtler input-format
discontinuities (e.g., morphological boundaries in inflected languages)
is an open question.

\emph{Ceiling accuracy.} The pre-registered discrimination hypotheses
(H2, H3, H5) were not evaluable due to ceiling accuracy. Arabic numeral
comparison is trivially easy for capable LLMs. A harder task (e.g.,
cross-format comparison: ``Which is larger, nine or 12?'') would enable
the full SDT analysis.

\emph{Control-150 leakage.} The control-150 condition showed small but
non-zero CP advantages (+0.012 to +0.042), suggesting possible leakage
of the 100-boundary effect. These values were far smaller than the
decade-100 effects (+0.162 to +0.476), but the control is not as clean a
null as control-15 for the decade-10 boundary.

\emph{Temperature domain.} Only one non-structural category boundary was
tested. ``Hot/cold'' is a relatively fuzzy, gradient boundary.
Additional domains with sharper linguistic categories (e.g., grammatical
number, animacy) would strengthen the cross-domain generalisation.

\emph{Scale.} The study tested models in the 3.8 to 9.2B parameter
range. Whether the same patterns hold at substantially larger scales
(70B+) is unknown.

\emph{Tokenisation mechanism.} The Supplementary Table S2 analysis
reveals that Llama-3's BPE vocabulary merges multi-digit numbers into
single tokens, eliminating the token-count discontinuity present in the
other four tokenisers. Yet Llama shows comparable CP geometry. This
confirms that character-count and lexical-form changes are sufficient,
and that token-count changes amplify but are not necessary. A
same-domain, different-tokenisation control (e.g., re-tokenising
identical stimuli with alternative BPE vocabularies) would provide a
stronger test.

\hypertarget{conclusion}{%
\subsubsection{4.9 Conclusion}\label{conclusion}}

Structural input-format discontinuities, including tokenisation
boundaries, digit-count transitions, and character-length changes, are
sufficient to produce a robust geometric signature consistent with
categorical perception in LLM hidden states, independently of explicit
semantic category knowledge. The geometric warping at category
boundaries is universal across architectures, present in pretrained
representations before instruction tuning, absent in domains that lack
structural input discontinuities, and causally grounded at early layers
where representations are still forming rather than at later layers
where the geometry reaches its peak legibility.

The dissociation between geometric CP (universal) and explicit
identification (architecture-dependent) demonstrates that
representational structure can outrun behavioural competence. For AI
interpretability, this cautions against equating the most legible
representation with the most functionally relevant one. For theories of
categorical perception more broadly, it provides a computational
existence proof that CP geometry can arise from the format of the input
representation, not only from category learning, in a system where both
the format and the learning are fully observable.

\hypertarget{acknowledgments}{%
\subsection{Acknowledgments}\label{acknowledgments}}

This research was conducted as part of the Classical Minds, Modern Machines independent research programme. No external funding was received.

\hypertarget{declaration-of-generative-ai-and-ai-assisted-technologies-in-the-manuscript-preparation-process}{%
\subsection{Declaration of Generative AI and AI-Assisted Technologies in
the Manuscript Preparation
Process}\label{declaration-of-generative-ai-and-ai-assisted-technologies-in-the-manuscript-preparation-process}}

During the preparation of this work the author used Claude (Anthropic)
in order to assist with code generation, figure production, and
manuscript preparation. After using this tool, the author reviewed and
edited the content as needed and takes full responsibility for the
content of the published article.

\hypertarget{data-availability}{%
\subsection{Data Availability}\label{data-availability}}

All code, stimuli, pre-registration, and analysis scripts are available
at https://github.com/synthiumjp/weber (directory
\texttt{m3\_pilot/}). Raw hidden-state extractions are available upon
request. Pre-registration:
https://osf.io/qrxf3.

\hypertarget{references}{%
\subsection{References}\label{references}}

Cacioli, J.-P. (2026a). Weber's Law in large language model hidden states.
Under review.

Cacioli, J.-P. (2026b). Do LLMs know what they know? Signal detection theory
meets metacognition. Under review.

Bonnasse-Gahot, L., \& Nadal, J.-P. (2022). Categorical perception: A
groundwork for deep learning. \emph{Neural Computation}, \emph{34}(2),
437--475.

Burns, E. M., \& Ward, W. D. (1978). Categorical perception ---
phenomenon or epiphenomenon: Evidence from experiments in the perception
of melodic musical intervals. \emph{Journal of the Acoustical Society of
America}, \emph{63}(2), 456--468.

Dehaene, S., Dupoux, E., \& Mehler, J. (1990). Is numerical comparison
digital? Analogical and symbolic effects in two-digit number comparison.
\emph{Journal of Experimental Psychology: Human Perception and
Performance}, \emph{16}(3), 626--641.

Dehaene, S., \& Mehler, J. (1992). Cross-linguistic regularities in the
frequency of number words. \emph{Cognition}, \emph{43}(1), 1--29.

de Wit, L., Alexander, D., Ekroll, V., \& Wagemans, J. (2016). Is
neuroimaging measuring information in the brain? \emph{Psychonomic
Bulletin \& Review}, \emph{23}(5), 1415--1428.

Elhage, N., Nanda, N., Olsson, C., Henighan, T., Joseph, N., Mann, B.,
Askell, A., Bai, Y., Chen, A., Conerly, T., DasSarma, N., Drain, D.,
Ganguli, D., Hatfield-Dodds, Z., Hernandez, D., Jones, A., Kernion, J.,
Lovitt, L., Ndousse, K., \ldots{} Olah, C. (2022). Toy models of
superposition. \emph{arXiv:2209.10652}.

Etcoff, N. L., \& Magee, J. J. (1992). Categorical perception of facial
expressions. \emph{Cognition}, \emph{44}(3), 227--240.

Goldstone, R. L. (1994). Influences of categorization on perceptual
discrimination. \emph{Journal of Experimental Psychology: General},
\emph{123}(2), 178--200.

Goldstone, R. L., \& Hendrickson, A. T. (2010). Categorical perception.
\emph{Wiley Interdisciplinary Reviews: Cognitive Science}, \emph{1}(1),
69--78.

Harnad, S. (Ed.). (1987). \emph{Categorical perception: The groundwork
of cognition}. Cambridge University Press.

Hinrichs, J. V., Yurko, D. S., \& Hu, J. M. (1981). Two-digit number
comparison: Use of place information. \emph{Journal of Experimental
Psychology: Human Perception and Performance}, \emph{7}(4), 890--901.

Liberman, A. M., Harris, K. S., Hoffman, H. S., \& Griffith, B. C.
(1957). The discrimination of speech sounds within and across phoneme
boundaries. \emph{Journal of Experimental Psychology}, \emph{54}(5),
358--368.

Massaro, D. W. (1987). Categorical partition: A fuzzy-logical model of
categorization behavior. In S. Harnad (Ed.), \emph{Categorical
perception: The groundwork of cognition} (pp.~254--283). Cambridge
University Press.

McMurray, B. (2022). The myth of categorical perception. \emph{Journal
of the Acoustical Society of America}, \emph{152}(6), 3819--3842.

Nuerk, H.-C., Weger, U., \& Willmes, K. (2001). Decade breaks in the
mental number line? Putting the tens and units back in different bins.
\emph{Cognition}, \emph{82}(1), B25--B33.

Nuerk, H.-C., Moeller, K., Klein, E., Willmes, K., \& Fischer, M. H.
(2011). Extending the mental number line: A review of multi-digit number
processing. \emph{Zeitschrift f\"{u}r Psychologie / Journal of Psychology},
\emph{219}(1), 3--22.

Park, K., Yun, S., Lee, J., \& Shin, J. (2024). The geometry of
categorical and hierarchical concepts in large language models.
\emph{arXiv:2406.01506}.

Park, K., Yun, S., Lee, J., \& Shin, J. (2025). The linear
representation of categorical and hierarchical concepts in LLMs. In
\emph{Proceedings of the International Conference on Learning
Representations (ICLR 2025)} (Oral).

Pollmann, T., \& Jansen, C. (1996). What round numbers tell us.
\emph{Cognition}, \emph{59}(2), 209--237.

Poltrock, S. E., \& Schwartz, D. R. (1984). Comparative judgements of
multi-digit numbers. \emph{Journal of Experimental Psychology: Learning,
Memory, and Cognition}, \emph{10}(1), 32--45.

Rogers, J. C., \& Davis, M. H. (2009). Categorical perception of speech
without stimulus repetition. In \emph{Proceedings of Interspeech 2009}
(pp.~376--379).

Shani, C., Marjieh, R., \& Griffiths, T. L. (2025). Compression in LLMs
mirrors human typicality gradients. \emph{arXiv:2505.17117}.

Shah, R., Marupudi, V., Koenen, R., Bhardwaj, K., \& Varma, S. (2023).
Numeric magnitude comparison effects in large language models. In
\emph{Findings of the Association for Computational Linguistics: ACL
2023} (pp.~6147--6161). Association for Computational Linguistics.

Winawer, J., Witthoft, N., Frank, M. C., Wu, L., Wade, A. R., \&
Boroditsky, L. (2007). Russian blues reveal effects of language on color
discrimination. \emph{Proceedings of the National Academy of Sciences},
\emph{104}(19), 7780--7785.

Zhu, F., Dai, D., \& Sui, Z. (2025). Language models encode the value of
numbers linearly. In \emph{Proceedings of the International Conference
on Computational Linguistics (COLING 2025)} (pp.~693--709).

\hypertarget{figure-legends}{%
\subsection{Supplementary Materials}\label{supplementary-materials}}

\hypertarget{supplementary-table-s2-tokeniser-encoding-of-boundary-relevant-numbers}{%
\subsubsection{Supplementary Table S2: Tokeniser Encoding of
Boundary-Relevant
Numbers}\label{supplementary-table-s2-tokeniser-encoding-of-boundary-relevant-numbers}}

Token counts for numbers spanning the decade-10 (9-\textgreater10) and
decade-100 (99-\textgreater100) boundaries. Three tokenisation patterns
emerge: (i) Llama-3 (instruct and base) uses merged BPE tokens for
multi-digit numbers, producing no token-count discontinuity at either
boundary; (ii) Gemma and Qwen use per-digit tokenisation with
single-digit numbers as one token, yielding a 1-\textgreater2 token jump
at the decade-10 boundary and 2-\textgreater3 at decade-100; (iii)
Mistral and Phi prepend a leading-space token and tokenise per-digit,
yielding 2-\textgreater3 and 3-\textgreater4 token jumps respectively.

\textbf{Decade-10 boundary:}

\begin{longtable}[]{@{}
  >{\raggedright\arraybackslash}p{(\columnwidth - 12\tabcolsep) * \real{0.1392}}
  >{\raggedright\arraybackslash}p{(\columnwidth - 12\tabcolsep) * \real{0.1392}}
  >{\raggedright\arraybackslash}p{(\columnwidth - 12\tabcolsep) * \real{0.1392}}
  >{\raggedright\arraybackslash}p{(\columnwidth - 12\tabcolsep) * \real{0.1392}}
  >{\raggedright\arraybackslash}p{(\columnwidth - 12\tabcolsep) * \real{0.1392}}
  >{\raggedright\arraybackslash}p{(\columnwidth - 12\tabcolsep) * \real{0.1392}}
  >{\raggedright\arraybackslash}p{(\columnwidth - 12\tabcolsep) * \real{0.1392}}@{}}
\toprule\noalign{}
\begin{minipage}[b]{\linewidth}\raggedright
Number
\end{minipage} & \begin{minipage}[b]{\linewidth}\raggedright
Digits
\end{minipage} & \begin{minipage}[b]{\linewidth}\raggedright
Llama-3
\end{minipage} & \begin{minipage}[b]{\linewidth}\raggedright
Mistral
\end{minipage} & \begin{minipage}[b]{\linewidth}\raggedright
Gemma
\end{minipage} & \begin{minipage}[b]{\linewidth}\raggedright
Qwen
\end{minipage} & \begin{minipage}[b]{\linewidth}\raggedright
Phi
\end{minipage} \\
\midrule\noalign{}
\endhead
\bottomrule\noalign{}
\endlastfoot
9 & 1 & 1 tok & 2 tok & 1 tok & 1 tok & 2 tok \\
10 & 2 & 1 tok & 3 tok & 2 tok & 2 tok & 3 tok \\
\textbf{$\Delta$ tokens} & & \textbf{0} & \textbf{+1} & \textbf{+1} &
\textbf{+1} & \textbf{+1} \\
\end{longtable}

\textbf{Decade-100 boundary:}

\begin{longtable}[]{@{}
  >{\raggedright\arraybackslash}p{(\columnwidth - 12\tabcolsep) * \real{0.1392}}
  >{\raggedright\arraybackslash}p{(\columnwidth - 12\tabcolsep) * \real{0.1392}}
  >{\raggedright\arraybackslash}p{(\columnwidth - 12\tabcolsep) * \real{0.1392}}
  >{\raggedright\arraybackslash}p{(\columnwidth - 12\tabcolsep) * \real{0.1392}}
  >{\raggedright\arraybackslash}p{(\columnwidth - 12\tabcolsep) * \real{0.1392}}
  >{\raggedright\arraybackslash}p{(\columnwidth - 12\tabcolsep) * \real{0.1392}}
  >{\raggedright\arraybackslash}p{(\columnwidth - 12\tabcolsep) * \real{0.1392}}@{}}
\toprule\noalign{}
\begin{minipage}[b]{\linewidth}\raggedright
Number
\end{minipage} & \begin{minipage}[b]{\linewidth}\raggedright
Digits
\end{minipage} & \begin{minipage}[b]{\linewidth}\raggedright
Llama-3
\end{minipage} & \begin{minipage}[b]{\linewidth}\raggedright
Mistral
\end{minipage} & \begin{minipage}[b]{\linewidth}\raggedright
Gemma
\end{minipage} & \begin{minipage}[b]{\linewidth}\raggedright
Qwen
\end{minipage} & \begin{minipage}[b]{\linewidth}\raggedright
Phi
\end{minipage} \\
\midrule\noalign{}
\endhead
\bottomrule\noalign{}
\endlastfoot
99 & 2 & 1 tok & 3 tok & 2 tok & 2 tok & 3 tok \\
100 & 3 & 1 tok & 4 tok & 3 tok & 3 tok & 4 tok \\
\textbf{$\Delta$ tokens} & & \textbf{0} & \textbf{+1} & \textbf{+1} &
\textbf{+1} & \textbf{+1} \\
\end{longtable}

\textbf{Full token IDs (decade-10 range):}

\begin{longtable}[]{@{}
  >{\raggedright\arraybackslash}p{(\columnwidth - 10\tabcolsep) * \real{0.1176}}
  >{\raggedright\arraybackslash}p{(\columnwidth - 10\tabcolsep) * \real{0.2118}}
  >{\raggedright\arraybackslash}p{(\columnwidth - 10\tabcolsep) * \real{0.1529}}
  >{\raggedright\arraybackslash}p{(\columnwidth - 10\tabcolsep) * \real{0.1529}}
  >{\raggedright\arraybackslash}p{(\columnwidth - 10\tabcolsep) * \real{0.1647}}
  >{\raggedright\arraybackslash}p{(\columnwidth - 10\tabcolsep) * \real{0.1765}}@{}}
\toprule\noalign{}
\begin{minipage}[b]{\linewidth}\raggedright
Number
\end{minipage} & \begin{minipage}[b]{\linewidth}\raggedright
Llama-3 (Instruct/Base)
\end{minipage} & \begin{minipage}[b]{\linewidth}\raggedright
Gemma-2-9B
\end{minipage} & \begin{minipage}[b]{\linewidth}\raggedright
Qwen2.5-7B
\end{minipage} & \begin{minipage}[b]{\linewidth}\raggedright
Mistral-7B
\end{minipage} & \begin{minipage}[b]{\linewidth}\raggedright
Phi-3.5-mini
\end{minipage} \\
\midrule\noalign{}
\endhead
\bottomrule\noalign{}
\endlastfoot
4 & \texttt{{[}19{]}} (1 tok) & \texttt{{[}235310{]}} (1 tok) &
\texttt{{[}19{]}} (1 tok) & \texttt{{[}\textunderscore{},\ 4{]}} (2 tok) &
\texttt{{[}\textunderscore{},\ 4{]}} (2 tok) \\
5 & \texttt{{[}20{]}} (1 tok) & \texttt{{[}235308{]}} (1 tok) &
\texttt{{[}20{]}} (1 tok) & \texttt{{[}\textunderscore{},\ 5{]}} (2 tok) &
\texttt{{[}\textunderscore{},\ 5{]}} (2 tok) \\
6 & \texttt{{[}21{]}} (1 tok) & \texttt{{[}235318{]}} (1 tok) &
\texttt{{[}21{]}} (1 tok) & \texttt{{[}\textunderscore{},\ 6{]}} (2 tok) &
\texttt{{[}\textunderscore{},\ 6{]}} (2 tok) \\
7 & \texttt{{[}22{]}} (1 tok) & \texttt{{[}235324{]}} (1 tok) &
\texttt{{[}22{]}} (1 tok) & \texttt{{[}\textunderscore{},\ 7{]}} (2 tok) &
\texttt{{[}\textunderscore{},\ 7{]}} (2 tok) \\
8 & \texttt{{[}23{]}} (1 tok) & \texttt{{[}235321{]}} (1 tok) &
\texttt{{[}23{]}} (1 tok) & \texttt{{[}\textunderscore{},\ 8{]}} (2 tok) &
\texttt{{[}\textunderscore{},\ 8{]}} (2 tok) \\
9 & \texttt{{[}24{]}} (1 tok) & \texttt{{[}235315{]}} (1 tok) &
\texttt{{[}24{]}} (1 tok) & \texttt{{[}\textunderscore{},\ 9{]}} (2 tok) &
\texttt{{[}\textunderscore{},\ 9{]}} (2 tok) \\
10 & \texttt{{[}605{]}} (1 tok) & \texttt{{[}1,\ 0{]}} (2 tok) &
\texttt{{[}1,\ 0{]}} (2 tok) & \texttt{{[}\textunderscore{},\ 1,\ 0{]}} (3 tok) &
\texttt{{[}\textunderscore{},\ 1,\ 0{]}} (3 tok) \\
11 & \texttt{{[}806{]}} (1 tok) & \texttt{{[}1,\ 1{]}} (2 tok) &
\texttt{{[}1,\ 1{]}} (2 tok) & \texttt{{[}\textunderscore{},\ 1,\ 1{]}} (3 tok) &
\texttt{{[}\textunderscore{},\ 1,\ 1{]}} (3 tok) \\
12 & \texttt{{[}717{]}} (1 tok) & \texttt{{[}1,\ 2{]}} (2 tok) &
\texttt{{[}1,\ 2{]}} (2 tok) & \texttt{{[}\textunderscore{},\ 1,\ 2{]}} (3 tok) &
\texttt{{[}\textunderscore{},\ 1,\ 2{]}} (3 tok) \\
15 & \texttt{{[}868{]}} (1 tok) & \texttt{{[}1,\ 5{]}} (2 tok) &
\texttt{{[}1,\ 5{]}} (2 tok) & \texttt{{[}\textunderscore{},\ 1,\ 5{]}} (3 tok) &
\texttt{{[}\textunderscore{},\ 1,\ 5{]}} (3 tok) \\
20 & \texttt{{[}508{]}} (1 tok) & \texttt{{[}2,\ 0{]}} (2 tok) &
\texttt{{[}2,\ 0{]}} (2 tok) & \texttt{{[}\textunderscore{},\ 2,\ 0{]}} (3 tok) &
\texttt{{[}\textunderscore{},\ 2,\ 0{]}} (3 tok) \\
\end{longtable}

Note: Llama-3 (both Instruct and Base) shares the same tokeniser; token
IDs are identical. The \texttt{\textunderscore{}} symbol denotes a leading-space token
prepended by Mistral and Phi tokenisers. Despite the absence of a
token-count discontinuity in Llama-3, all models show comparable CP
geometry at the decade-10 boundary (Table 1), indicating that
character-count and lexical-form changes are sufficient to produce the
effect.

\emph{Draft generated April 2026. Correspondence: {[}anonymised for
review{]}.}

\end{document}